# I3rab: A New Arabic Dependency Treebank Based on Arabic Grammatical Theory

Dana Halabi

Computer Science Department, Princess Sumaya University for Technology, Amman, Jordan, d.halabi@psut.edu.jo

Ebaa Fayyoumi

Computer Science Department, Hashemite University, Zarqa, Jordan, enfayyoumi@hu.edu.jo

Arafat Awajan

Computer Science Department, Princess Sumaya University for Technology, Amman, Jordan, awajan@psut.edu.jo

## ABSTRACT

Treebanks are valuable linguistic resources that include the syntactic structure of a language sentence in addition to POS-tags and morphological features. They are mainly utilized in modeling statistical parsers. Although the statistical natural language parser has recently become more accurate for languages such as English, those for the Arabic language still have low accuracy.

The purpose of this paper is to construct a new Arabic dependency treebank based on the traditional Arabic grammatical theory and the characteristics of the Arabic language, to investigate their effects on the accuracy of statistical parsers. The proposed Arabic dependency treebank, called I3rab, contrasts with existing Arabic dependency treebanks in two main concepts. The first concept is the approach of determining the main word of the sentence, and the second concept is the representation of the joined and covert pronouns.

To evaluate I3rab, we compared its performance against a subset of Prague Arabic Dependency Treebank that shares a comparable level of details. The conducted experiments show that the percentage improvement reached up to 7.5% in UAS and 18.8% in LAS.

## 1 Introduction

Treebanks are annotated corpora that serve as valuable resources in many data-driven Natural Language Processing (NLP) applications. Typically, sentences in a treebank are annotated with part of speech (POS), morphological features, and syntactic structure (Frank, Zaenen, and Hinrichs 2012; Volk et al. 2005). Treebanks are mainly used for modeling statistical parsers (Kübler, McDonald, and Nivre 2009). Furthermore, they are also used in other NLP applications such as question-answering (Li and Xu 2016; Comas, Turmo, and Márquez 2010; Bouma et al. 2005), machine translation (Galley and Manning 2009; Katz-Brown et al. 2011; Ambati 2008), evaluation of machine translation (McCaffery and Nederhof 2016; Yu et al. 2015; Owczarzak, Van Genabith, and Way 2007) and information retrieval (Gillenwater et al. 2013).

There are many grammatical formalisms to represent the syntax structure. The most commonly used are phrase structure and dependency structure (Xia and Palmer 2001). The first constructed treebank was a phrase structure treebank for English. It was the first large-scale syntactic annotation treebank developed and distributed by the Linguistic Data Consortium (LDC) (Marcus, Santorini, and Marcinkiewicz 2006). It has primarily been used to develop the statistical parser model for the English language (Collins 2003). The success of using treebanks in the English language inspired researchers to follow the same approach to develop phrase-structure treebanks for various languages such as Catalan (2004) (Civit, Bufí, and Valverde 2004), Spanish (2004) (Civit and Martí 2004), Chinese (1998) (Xue et al. 2013), Arabic (2003) (Maamouri et al. 2004), Hebrew (2001) (Sima'an et al. 2001) and Korean (2002) (Han et al. 2002).

The first dependency treebank was the Prague Dependency Treebank (PDT) (Böhmová et al. 2003) developed for the Czech language. The dependency syntax of the PDT was strongly influenced by the functional generative description theory, which considers the verb as the main word in the sentence, regardless of its position. In addition to the Czech language, a dependency treebank approach has been adopted for other languages, such as Arabic, Basque, Catalan, Chinese, Czech, English, Greek, Hungarian, Italian and Turkish. More attention has been given to dependency treebanks through the Conference on Natural Language Learning (CoNLL) 2007 (Nivre et al. 2007a), dedicated to dependency parsing.

Several treebanks have been developed for Arabic. The most important treebanks are the Penn Arabic Treebank (PAT) (Maamouri et al. 2009, 2004), Prague Arabic Dependency Treebank (PADT) (Smrz, Bielicky, and Hajic 2008), Columbia Arabic Treebanks (CATiB) (Habash and Roth 2009), Classical Arabic Treebank (Dukes and Buckwalter 2010) and US Army Research Laboratory (ARL) Arabic Dependency Treebank (AADT) (Tratz 2016). PAT (Maamouri et al. 2009, 2004) was the first treebank developed for Modern Standard Arabic (MSA). It is a phrase structure treebank based on the Penn English Treebank (Marcus, Santorini, and Marcinkiewicz 2006), and it uses the same tagset of the Penn English Treebank to annotate the phrase structure of PAT. Initially, it used more than 400 POS tags, but the number was reduced to 36 POS tags during training and testing parsers (Kulick, Gabbard, and Marcus 2006).

PADT (Smrz, Bielicky, and Hajic 2008) is a dependency treebank for the same text sources in PAT. The dependency syntax of PADT is strongly influenced by the PDT. The authors in (Smrz, Bielicky, and Hajic 2008) have argued that the Arabic and Czech languages are rich in inflection and share the free word order property. The verb is considered the main word in the sentence regardless of its position in the sentence. However, the PADT team manually treated the features of the Arabic language that could not be handled in the same way as the Czech language.

Another dependency treebank for MSA is CATiB (Habash and Roth 2009). The dependency labels of CATiB were inspired by traditional Arabic grammatical theory, but this treebank uses a small subset of the full traditional syntactic roles. It has only six POS tags and eight relation labels for dependency links. Parsers trained against this treebank can accelerate the development of new treebanks but with limitations in linguistic richness. Although the dependency labels of the CATiB treebank were based on traditional Arabic grammar, it was inspired by PDT in considering the verb as the main word in the sentence regardless of its position in the sentence.

The Classical Arabic Treebank (Dukes and Buckwalter 2010) is an annotated corpus specialized for the text of the Holy Quran. The Holy Quran is a major religious text that is considered to contain unique and challenging language. The Holy Quran is a collection of 114 ordered chapters (سور, suar), each with a number of ordered verses (آيات, ayat). The syntactic tree for a verse is represented in a hybrid dependency-constituency phrase structure model. The syntactic tree depends on the traditional Arabic grammar exposed in the well-known book (إعراب القرآن الكريم, 'irabu alqurani alkarim). The author in (Dukes and Habash 2011; Dukes 2015) have argued that this hybrid representation is sufficiently flexible to represent all aspects of the syntax in the Holy Quran.

The newest dependency treebank for MSA is AADT (Tratz 2016). It was derived from existing Arabic treebanks distributed by LDC, by using constituent-to-dependency conversion tools. The dependency scheme consists of a total of 35 labels. Many of these are similar to those of Stanford's basic dependency scheme for English. Similarly, to CATiB, this treebank is based on the idea that the verb is the main word in the sentence.

Arabic statistical parsers have been developed on the basis of phrase structure, e.g., PAT (Kulick, Gabbard, and Marcus 2006), and dependency-structure, e.g., PADT (Nivre et al. 2007a; Smrz, Bielicky, and Hajic 2008). The authors of (Kulick, Gabbard, and Marcus 2006) reported that the accuracy of Arabic parsing obtained by using PAT version 1 to train the Bikel parser (Bikel 2004) had an F1-score of 74%, which was considered a low score relative to the 88% F1-score for English on a comparably sized corpus.

CoNLL shared task 2007 was devoted to dependency parsing. It involved developing and evaluating different state-of-the-art parsers for ten languages. The dependency parsers generated for the Arabic language were based on the PADT (Nivre et al. 2007a; Smrz, Bielicky, and Hajic 2008). The best performance for the dependency parsing of Arabic was found to be 76.52% for the Labeled Attachment Score (LAS). The LAS values were separated into three classes: low, medium and high. The Arabic language received a low score, whereas the Czech language received a medium score of 80.2%, and the English language received a high score of 89.6% (Nivre et al. 2007a).

In spite of having good performance by using both syntactic representation in English and Czech parsers, both syntactic representations result in low performance for MSA parsers. This low accuracy indicates the need to investigate the reasons underlying the low performance of the various parsers. In general, the results obtained by a parser are related to two main issues: (1) the quality of the linguistic resources (treebanks) used in modeling the statistical parsers, and (2) the approaches and algorithms used in developing the parsers. This paper investigates and focuses on the quality of the linguistic resources (treebanks) to improve the parser performance.

In general, two factors affect the quality of treebanks. The first factor is the level and quality of annotation. The second factor is the concepts and theories involved in analyzing the structure of a sentence and mapping the structure into a phrase structure or dependency structure within the treebank. In the case of Arabic, for both treebanks, the first factor was covered in an accepted way. Both treebanks use deep analysis of sentences including POS tags, morphological analysis, diacritization and lemmas, beyond syntactic annotation (Maamouri et al. 2004; Smrz, Bielicky, and Hajic 2008). The annotation process has been performed and rechecked manually to increase the quality of the annotation information (Habash and Roth 2009). However, for the second factor, the PAT was strongly inspired by English treebank, whereas the PADT was strongly influenced by the PDT developed for the Czech language and inspired by the functional generative description theory in considering the verb as the main word in the sentence regardless of its position. Therefore, the concepts used in developing the existing MSA treebanks were inspired by the characteristics of other languages, mainly English and Czech (Ryding 2005), (Maamouri et al. 2004), (Smrz, Šnaidauf, and Zemánek 2002).

Consequently, there is a need for a new Arabic treebank constructed according to linguistic and grammatical theories covering the Arabic features, and simultaneously compatible with the concepts and rules of constructing treebanks. The first step to addressing this issue is determining the most appropriate grammatical formalisms that the Arabic linguistic and grammatical theories should coincide with. In this paper, we selected the dependency structure on the basis of the substantial attention that has been paid to dependency-structure treebanks in the past two decades. The reason for this decision was the usefulness of bi-lexical relations between individual words (head and modifier words) in solving different ambiguity problems in POS and parsing tasks (Nivre 2005; Kübler, McDonald, and Nivre 2009). Moreover, the Arabic language is a Semitic language that is highly inflectional and has rich morphological features (Al-Sughaiyer and Al-Kharashi 2004). In addition, it is considered to have relatively flexible word order (Ryding 2005). Bharati (Bharati et al. 1995) has suggested that free word order and rich morphological languages can be handled better by using a dependency based rather than phrase-structure based framework.

The Arabic language, like other languages, has several linguistic and grammatical theories dedicated to describing its features and characteristics (Alosh 2005; Owens 1988, 1990). This paper is based on the traditional Arabic grammatical theory called I‘rab. One of the main concepts of I‘rab is the categorization of Arabic sentences into two categories - verbal sentence or nominal sentence - depending on the type of the first word in the sentence. Another main concept is that each verb should have a subject, and this subject should follow the verb and cannot precede it. I‘rab considers all forms of the verb's subject, whether it is a nominative noun or independent, joined or covert pronoun (Alosh 2005; Owens 1988).

The objective of this paper is to construct the I3rab1 treebank, a new pilot dependency treebank for MSA that is based on I‘rab theory. I3rab contrasts with the existing MSA treebanks in two main aspects. First, I3rab is completely different from the exiting dependency MSA treebanks in how the main word in the sentence is determined. In the existing MSA dependency treebanks, for the sentence that has a verb, the verb is considered as the main word regardless of its position. In contrast, I3rab determines the first word in the sentence as the main word. From this perspective, in an Arabic sentence that starts with a verb, both I3rab and the existing MSA dependency treebanks consider the verb as the main word and have a mostly similar dependency structure representing the sentence. However, if a sentence starts with a noun, I3rab considers the noun as the main word for building its dependency structure, regardless of whether a verb exists in the sentence. If the sentence has no verb, then these treebanks consider the predicate (comment) as the main word (Smrz, Šnaidauf, and Zemánek 2002; Habash, Faraj, and Roth 2009). The second difference is the approach of handling the subject pronoun, whether it is a joined or covert pronoun. I3rab has explicit presentation for subject pronouns, whereas existing MSA dependency treebanks have no explicit representation for these pronouns. This paper demonstrates how I‘rab can be used to construct a consistent dependency structure.

[1] I3rab is derived from the English transliteration (I`rab) of the Arabic word (اعراب). In general, Arabic speakers replace the Arabic letter (ع) with the number (3) in the Arabization text that is usually used in chats and WhatsApp conversations.

This paper is organized as follows. The theoretical concepts of the proposed I3rab dependency treebank are covered in Section 2. The proposed I3rab dependency treebank is covered in Section 3. The implementation and experimental results are discussed in Section 4. Finally, the conclusion and future work are described in Section 5.

# 2 Theoretical concepts of the proposed I3rab dependency treebank

I‘rab is derived from a general well-known Arabic traditional theory called “The Theory of Al Aamil” (نظرية العامل, nazariatu aleamili). It appeared in the second Hijree century (approximately 900 A.D.). The main aim was to aid foreigners in learning the Arabic language in addition to studying and understanding the Holy Quran. The Theory of AL Aamil consists of three main components: governance (العمل, alamalu), governor (العامل, alamilu) and governed (المعمول, almamulu).

In this section, the main features of the Arabic language related to I‘rab will be covered and followed by the main essential principles of I‘rab. Then the main concepts of I‘rab and its relationship to dependency grammar will be explained.

## 2.1 Arabic features

The Arabic language is a highly inflectional and rich morphological language and therefore exhibits many complexities that pose interesting challenges for NLP tasks related to disambiguation, whether these tasks are preprocessing tasks, such as segmentation and tokenization processes and POS tagging, or they mediate processes such as parsing (Attia and Somers 2008; Awajan 2015). Disambiguation is a major challenge that influences the quality of the results of Arabic NLP tasks. The Arabic language and MAS texts have many features that increase the degree of ambiguity.

The Arabic language is a highly inflectional language, thus making morphological analysis complicated (Attia and Somers 2008; Awajan 2007). Arabic words are divided into derivative and non-derivative words. Derivative words are produced by applying a wide range of standard patterns producing different surface forms for the word (Awajan 2016). These patterns define the features of the word, such as gender, number and tense for the verb (Awajan 2015, 2007). For example, the word (مزارع, mazarieun, farmer) is a masculine, singular word, whereas the word (مزارعون, muzarieuna, farmers) is a masculine, plural word. The part (ون, una) is used to indicate that the word is a masculine plural noun.

Arabic is a clitic language, in which a single word may actually be a complete sentence or phrase (Awajan 2007; Attia 2007; Alotaiby, Foda, and Alkharashi 2010). In English, for example, the word (didn’t) is divided into two tokens: the first token is (did), and the second token is (n’t). The orthographic mark (’) is an indication of a clitic in the English language, but in the Arabic language, clitics are combined with Arabic words as pro-clitics or enclitics without orthographic marks (Awajan 2015, 2007; Alotaiby, Foda, and Alkharashi 2010). For example, the Arabic word (لبيته, li bayti hi, to his house) is divided into three tokens: the first token is (لـ, li, to), a prepositional particle; the second token is (بيت, bayti, house), a noun; and the third token is (ـه, hi, his), an object pronoun. Another example is the verb (يزرع, yazraeuna, he plants); it has no clitics joined to it, but when a set of people are being referred to (يزرعون, yazraeuna, they plant), it has the part (ون, una), which represents a subject joined pronoun. In another example, in the words (مزارعون, muzarieuna, farmers) and (يزرعون, yazraeuna, they plant), the first word one is a noun, and the part (ون, una) represents the word's morphological features, whereas the second word is a verb, and the part (ون, una, they (plural)) represents a subject joined pronoun attached to a verb.

In addition, the Arabic language is a pro-drop language (Farghaly and Shaalan 2009). The pro-drop is a syntactic phenomenon in which the subject is omitted, owing to the ability to understand the subject from the morphological features of the verb (Farghaly and Shaalan 2009; Chomsky 1993). For example, in the sentence (محمد فهم الدرس, muhamadun fahima aldarsa, Mohammed understood the lesson) the subject pronoun (هو, huwa, him) is dropped, because it is understood from the context of the sentence that this dropped pronoun refers to (محمد, muhamadun, Mohammed). In computational linguistics related to the Arabic language, there is no agreement on how to address this un-lexicalized item. In the Arabic language, the PAT represents the pro-drop as an empty category, whereas in the PADT, it is ignored, on the basis of the concept that the subject is conjugated to the verb as part of its inflection (Ryding 2005; Hajic et al. 2004).

Arabic has a flexible word order. The most common structure of the Arabic sentence is a verb followed by a subject followed by an object (Attia and Somers 2008; Ryding 2005). However, in some cases, the writer chooses to start with a noun rather than a verb (Ryding 2005). Another issue related to structure is that the Arabic language can have sentences that have no verb, called pure nominal sentences, in which both the subject and predicate are two nominative nouns (Ryding 2005) such as (الشمس مشرقة, alshamsu mushriqatun, the sun is shiny).

## 2.2 The main essential principles of I`rab

Herein, the main essential principles of I`rab involved in this paper will be discussed.

1. The Arabic word classes

In the Arabic language, words can be categorized into three main classes: noun, verb or particle.

i. Noun: a noun is a word that conveys meaning on its own and does not have a tense. It may be a common noun (كتاب, kitabun, book), proper noun (أحمد, ahmadu, Ahmad), adjective (جميل, gamilun, beautiful), demonstrative (هذا, hada, this), personal pronoun (هي, hiya, she), relative pronoun (الذي, alladi, which) or numeral (خمسون, khamsuna, fifty) (Alosh 2005; Owens 1988).
ii. Verb: a verb is a word that describes an event or action being done. It conveys a meaning on its own, and it should have a tense. A verb can have three tenses: perfect (ماضي, madi), imperfect (مضارع, mudarieu) and imperative (أمر, amrun). In addition, the verb may be strong (تام, tamu) or defective (ناقص, naqisu). Most verbs are strong verbs, such as the perfect verb (كتب, kataba, he wrote), imperfect verb (يكتب, yaktubu, he writes) and imperative verb (اكتب, uktub, imperative order: write). The defective verb is part of abolishers (النواسخ, alnawasikhu) called auxiliary verbs (Alosh 2005).
iii. Particle: a particle is a word that does not convey meaning on its own and is combined with another word to provide meaning. Particles have a wide range of types (Alosh 2005; Owens 1988), for example, preposition particles (من, min, from), interrogative particles (هل, hal, whether), accusative particles (إنَّ, 'inna, indeed), coordinating particles (و, wa, and) and many others.

2. The sentence in I`rab

I`rab considers the sentence as the basic unit of analysis (Alosh 2005; Owens 1988). Sentences are categorized into two types, the verbal and the nominal sentences, depending on the class of the first word in the sentence.

i. The verbal sentence is a sentence starting with a strong verb. A sentence that starts with a defective verb is considered as a nominal sentence (Alosh 2005). A verbal sentence is minimally composed of two words: a verb followed by a noun (often called an agent). The agent may be a single nominative noun, pronoun or noun phrase. Each verb should have an agent, but the existence of the object is limited to the transitive verb only. For example, in the sentence (ينام الطفل, yanamu altiflu, the child sleeps), the verb is (ينام, yanamu, sleeps), and the agent is (الطفل, altiflu, the child). If the verb in the sentence is preceded by particles such as conjunctions (و, wa, and), accusative (لن, lan, will not), or jussive (لم, lam, did not) particles, the sentence is considered a verbal sentence. If the verb is preceded by an adverbial element (preposition phrase or adverbial phrase), it is considered verbal sentence. For example, (لم ينم الطفل باكراً, lam yanm altiflu bakraan, the child did not sleep early) and (في الصباح يذهب الأب إلى العمل, fi alsabahi yadhhabu al'abu 'iilaa aleamali, In the morning the father goes to work) are verbal sentences (Alosh 2005; Owens 1988).

ii. A nominal sentence is a sentence starting with a noun. It is minimally composed of two nominative successive nouns (noun + noun). Typically, the first noun is called a topic (مبتدأ, mubtadaun), and the second is called a predicate (خبر, khabarun). The topic can be a single nominative noun, pronoun or noun phrase (Alosh 2005; Owens 1988). For example, in the sentence (الشمس مشرقة, alshamsu mushriqatun, the sun is shiny.), the topic is (الشمس, alshamsu, the sun), and the predicate is (مشرقة, mushriqatun, shiny). The predicate can be in one of three cases (Alosh 2005; Owens 1988):
   a. A single nominative noun, for example is (مشرقة, mushriqatun, shiny) in the sentence (الشمس مشرقة, alshamsu mushriqatun, the sun is shiny.)
   b. A sentence, either a nominal sentence or a verbal sentence. In (الفتاة شعرها طويل., alfatatu shaeruha tawilun, the girl has long hair), the phrase (شعرها طويل, shaeruha tawilun, has long hair) is an entire nominal sentence that has a role as a predicate. In (محمد يقرأ الكتاب., muhamadun yaqrau alkitaba, Mohammed reads the book), the phrase (يقرأ الكتاب, yaqrau alkitaba, reads the book) is an entire verbal sentence that has the role of a predicate.
   c. An adverbial element, either a prepositional phrase or an adverbial phrase. In the sentence (العصفور في القفص., aleasfuru fi alqafsi, the bird the cage), the prepositional phrase (في القفص, fi alqafsi, in the cage) has a role as a predicate for the topic (العصفور, aleasfuru, the bird). In the sentence (الكتاب فوق الطاولة., alkitabu fawqa alttawilati, the book is above the table), the adverbial phrase (فوق الطاولة, fawqa alttawilati, is over the table) has a role as a predicate for the topic (الكتاب, alkitabu, the book).

The nominal sentence could be preceded by abolishers, and in this case it is still considered as a nominal sentence. An abolisher is a tool that introduces the nominal sentence and affects the syntax and the semantics of the sentence. In I'rab, there are mainly two groups of abolishers, Inn-it-sister (إن وأخواتها, 'inna wa 'akhawatiha) and Kana-its-sister (كان وأخواتها, kana wa 'akhawatiha). From the syntactic perspective, an Inn-its-sister is a set of particles that precedes the nominal sentence. These particles change the case of the topic to accusative and keep the predicate in nominative case. In contrast, a Kana-its-sister is a set of defective verbs that precede the nominal sentence. These verbs keep the topic in nominative case and change the case of the predicate to accusative case. From a semantic perspective, an abolisher provides extra information or sometime changes the meaning. For example, the abolisher (كان, kana, was) in the sentence (كان اليوم رائعا, kana alyawmu rayieaan, the day was wonderful) indicates that the event happened in the past (Alosh 2005; Owens1988).

3. Pronouns in the Arabic language

A pronoun is a word that replaces a noun. Therefore, a pronoun and a noun are functionally equivalent. In the Arabic language, pronouns have a major role in the syntactic and semantic analysis of linguistic structures. They can be a subject, object or possessive pronoun. One important role is that in a nominal sentence, if the predicate is an entire sentence that is either nominal or verbal, it should have a pronoun (joined or covert) that refers to the main topic of the original nominal sentence (Alosh 2005; Owens 1988).

Pronouns show differences in their morphological features, such as person (first, second or third), number (singular, dual or plural) and gender (masculine or feminine). In Arabic, there are three types of pronouns: independent personal pronouns, joined pronouns and covert pronouns. The independent personal and joined pronouns are considered overt pronouns, because they are presented explicitly in the sentence, either as individual words for independent pronouns or attached to another word for joined pronouns (Alosh 2005; Owens 1988).

i. Independent personal pronouns

The independent personal pronouns in Arabic language can take the roles of either nominative nouns or accusative nouns. The independent personal pronouns that take the roles of nominative nouns are (أنا, 'ana, me), (أنتَ, anta, you (male, singular)), (أنتِ, anti, you (female, singular)), (أنتما, antuma, you (dual)), (أنتم, antum, you (male, plural)), (أنتن, antunna, you (female, plural)), (هو, huwa, he), (هي, hiya, she), (هما, huma, they (dual)), (هم, hum, they (male, plural)) and (هن, hunna, they (male, plural)). These pronouns have positions, such as the topic for a nominal sentence, agent for an active verb and deputy agent for a passive verb (Alosh 2005; Owens 1988). Further details can be founded in Appendix A. The independent accusative nouns are rare in MSA, but they have certain uses in classic Arabic (Ryding 2005).

ii. Joined personal pronouns

The joined personal pronouns in the Arabic language can take the roles of nominative nouns, accusative nouns and genitive nouns (Alosh 2005; Owens 1988).

The joined personal pronouns that take the roles of nominative nouns are five pronouns (ان, `alifu alaithnayni, they (dual) or you (dual)), (ون, wawu aljamaeati, they (male, plural) or you (male, plural)), (ن, nunu alniswati, they (female, plural) or you (female, plural)), (نا, na, we), (ت, ta'u alfaeili, I and you (male, female, singular)) and (ي, ya'u almukhatibati, you (female, singular)). These pronouns are joined only to verbs that are either strong or defective. They have the roles of agent for an active verb, deputy agent for a passive verb or subject for a Kana-its-sister. For example, in the sentence (أنتما تكتبان المقالة., 'antuma tuktubani almuqalata, you are writing the article), the term (تكتبان, tuktubani, are writing) according to the theory will be segmented into the verb (تكتب, tuktubu, writes) and the joined pronoun (ان). Further details can be founded in Appendix A.

The joined personal pronouns that take the roles of accusative or genitive nouns are four pronouns. The first pronoun is (هاء الغائب, ha'u alghayibi). This pronoun has five forms (ـه, ها, هما, هم, هن) that correspond to (he, she, they (dual), they (male, plural), they (female, plural)). The second pronoun is (ي, ya'u almutakalimi, me or my). The third pronoun is (كاف الخطاب, kafu alkhitabi). This pronoun has four forms (ك, كما, كم, كن) that correspond to (you (male, female, singular, dual, plural)). The last pronoun is (نا, us or we). These pronouns take the roles of accusative nouns when joined to verbs or Inna-its-sisters. They serve as an object of the verb or a subject of Inna (Alosh 2005; Ryding 2005). For example, in the sentence (المقالة كتبها الطالب., almaqalatu katabaha ataalibu, the article was written by the student.), the term (كتبها, katabaha) is segmented to the verb (كتب, kataba, wrote), and (ها) that is a joined personal pronoun acting as an object of the verb (Alosh 2005; Ryding 2005). In contrast, these pronouns take the roles of genitive nouns when joined to nouns or preposition particles. They indicate possession (possessive pronouns) when joined to a noun or the object of the preposition when joined to preposition particles. For example, in the sentence ( الفتاة شعرها طويل, alfatatu shaeruha tawilun, the girl has long hair), the term (شعرها, shaeruha, her hair) is segmented to the noun (شعر, shaeru, hair) and (ها) which is a joined personal pronoun indicating possession that refers to the topic (الفتاة, alfatatu, the girl). Further details can be founded in Appendix B.

iii. Covert pronouns

Arabic is a pro-drop language. Pro-drop is a syntax feature wherein the subject can be omitted, because it can be determined from the context of the sentence. The pro-drop is a form of ellipsis. In Arabic, the ellipsis is known as (الحذف, hadhf, deletion). The deletion can be for grammatical or semantic reasons. The process of deleted items interpretation is known as taqdîr (determine or surmise).

There are five covert (drooped, absent) personal pronouns in Arabic: (أنا, ana, I), (نحن, nahnu, we), (هو, huwa, he), (هي, hiya, she) and (أنت, `anta, you (male, singular)). These pronouns take the roles of nominative nouns, such as the role of agent for an active verb, deputy agent for a passive verb or subject for a Kana-its-sister. For example, in the sentence (محمد يقرأ الكتاب, muhamadun yaqrau alkitaba, Mohammed reads the book), the verb (يقرأ, yaqrau, reads) according to the theory has a dropped agent determined by the third person singular masculine pronoun (هو, huwa, he). Appendix A lists the covert pronouns, their morphological features and examples of sentences that have covert pronouns as agents for perfect, imperfect and imperative verbs. These pronounces are indicated by (*) in Appendix A (Alosh 2005; Owens 1988).

## 2.3 The I`rab theory

In Arabic language, the basic unit of analysis is the sentence (Owens 1988), which consists of meaningful words. Each word has a specific role in the sentence determined by syntactical rules, which is used to build a coherent structure (Owens 1988).

1. The main concepts

I`rab considers an Arabic sentence is made up of three main components: governance, governor and governed. The governor word is linked to the governed word on the basis of the principle that the governor word has the power to affect the governed word in some manner, and to determine its role in the sentence (Alosh 2005; Owens 1988). This theory maps the Arabic sentence as a set of dependency relations between words. The reason for this is that the governor is associated with the head in dependency grammar, and the governed is associated with the modifier (Alosh 2005; Owens 1988). In other words, the governor governs (do, operate) the governed and can change its case (for nouns) or mood (for verbs) according to its function. The noun has three cases: nominative, accusative and genitive. For the verb, only imperfect verbs show mood inflection. These mood inflections are indicative, subjunctive and jussive. Linguistically, applying the process of this theory on Arabic sentence is known as (`irab) (Alosh 2005; Owens 1988). For example, in the verbal sentence (يأكل الرجل السمك, yakulu alrajulu alsamaka, the man eats the fish), the verb (يأكل, yakulu, eats) is the main word in the sentence and plays a predominant role in the sentence governing the other two words, (الرجل, alrajulu, the man) as an agent and (السمك, alsamaka, the fish) as an object (Owens 1988). Consequently, both the agent and the object are governed by the main word (verb). Table 1 shows the details of the grammatical analysis of the sentence (يأكل الرجل السمك, yakulu alrajulu alsamaka, the man eats the fish). Below, we will use the term i`rab starting with a small "i" to indicate the grammatical analysis of a sentence.

**Table 1: The i'rab of the sentence (يأكل الرجل السمك, the man eats the fish)**

| Word | Word role (Arabic) | Word role (English) |
|---|---|---|
| يأكل | فعل مضارع مرفوع | Inductive imperfect verb |
| الرجل | فاعل مرفوع | Nominative noun in the role of the agent of the verb (يأكل) |
| السمك | مفعول به منصوب | Successive noun in the role of the object of the verb (يأكل) |

Preceding the above sentence with a jussive particle (لم, lam, not) that indicates negation would change the mood of the imperfect verb from the inductive to jussive mode. Table 2 shows the i`rab of the sentence (لم يأكل الرجل السمك, lam yakul alrajulu alsamaka, the man did not eat the fish). The existence of the jussive particle makes it the main word that governs the imperfect verb (يأكل, yakul, eats). There is no change in the roles of the other two words (الرجل, alrajulu, the man) and (السمك, alsamaka, the fish).

### Table 2: The i‘rab of the sentence (لم يأكل الرجل السمك, the man did not eat the fish)

| Word | Word role (Arabic) | Word role (English) |
|---|---|---|
| لم | حرف جزم | Jussive particle |
| يأكل | فعل مضارع مجزوم | Jussive imperfect verb |
| الرجل | فاعل مرفوع | Nominative noun in the role of the agent of the verb (يأكل) |
| السمك | مفعول به منصوب | Successive noun in the role of the object of the verb (يأكل) |

Thus, an imperfect verb is governed by the jussive particle, whereas the agent and the object are still governed by the imperfect verb. Figures 1 shows parsing trees related to two verbal sentences, respectively. The (S) symbol in the tree represents the root of the tree, the main word in the sentence is the unique child of S, and the relations between each parent and its children are labeled by the role (function) of the child word in the sentence.

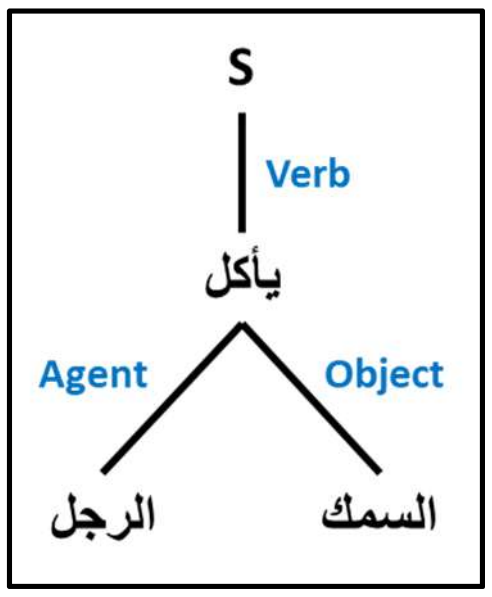

**(a) The tree of the sentence (يأكل الرجل السمك, the man eats the fish)**

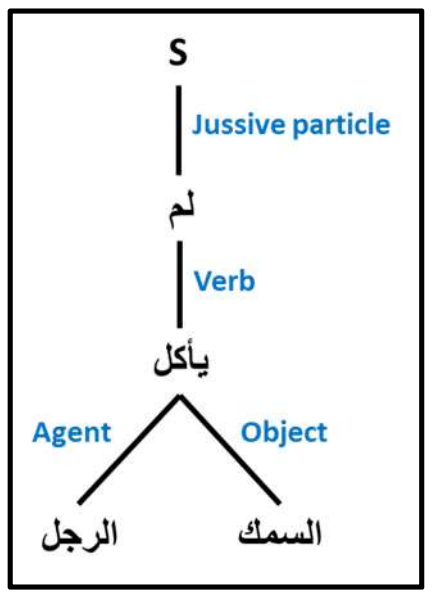

**(b) The tree of the sentence (لم يأكل الرجل السمك, the man did not eat the fish)**

**Figure 1: The syntactic structure for Arabic verbal sentence**

In the case of the two nominal sentences the sentence (الشمس مشرقة., alshamsu mushriqatun, the sun is shiny.) and (كانت الشمس مشرقة., kanat alshamsu mushriqatan, the sun was shining.), both sentences are similar, but the second sentence starts with (كانت, kanat, was), one of the Kana-its-sisters that indicates past tense. The i`rab of the two sentences are shown in Tables 3 and 4, respectively.

### Table 3: The i‘rab of the sentence (الشمس مشرقة., the sun is shiny.)

| Word | Word role (Arabic) | Word role (English) |
|---|---|---|
| الشمس | مبتدأ مرفوع | Nominative noun in the role of the topic |
| مشرقة | خبر المبتدأ مرفوع | Nominative noun in the role of the predicate of the topic (الشمس) |

**Table 4: The i‘rab of the sentence (كانت الشمس مشرقة., The sun was shining.)**

| Word | Word role (Arabic) | Word role (English) |
|---|---|---|
| كانت | فعل ماض ناقص | Defective perfect verb |
| الشمس | اسم كان مرفوع | Nominative noun in the role of the topic of (كانت) |
| مشرقة | خبر كان منصوب | Accusative noun in the role of the predicate of (كانت) |

In the first sentence, the topic (الشمس, alshamsu, the sun) is considered the main word that governs the predicate (مشرقة, mushriqatun, shiny) (Alotaiby, Foda, and Alkharashi 2010). In the second sentence, the abolisher (كانت, kanat, was) becomes the main word in the sentence and governs both the topic and the predicate. The tree representations of the above two nominal sentences are illustrated in Figure 2.

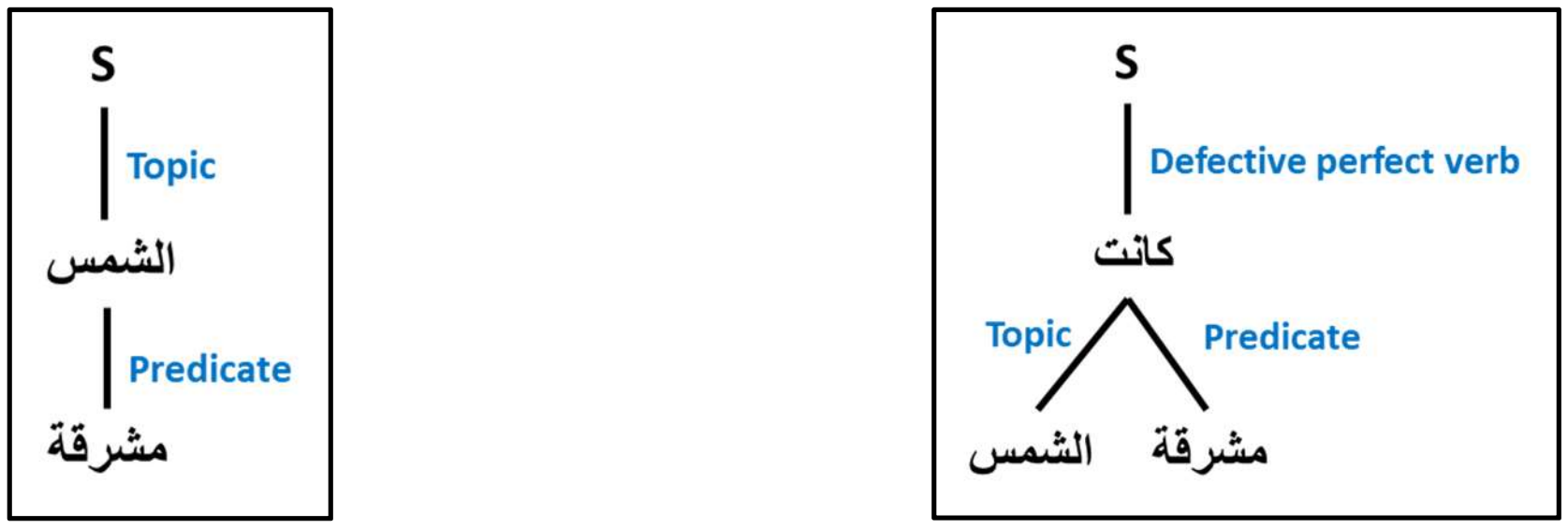

**(a) The tree of the sentence (الشمس مشرقة., the sun is shiny.)**

**(b) The tree of the sentence (كانت الشمس مشرقة., the sun was shining.)**

**Figure 2: The syntactic structure for Arabic nominal sentence**

2. The scope of the governor

The type of the sentence (nominal or verbal) enables understanding of its structure and consequently determines the main word in the sentence, the governor(s) and the governed word(s). Each governor has a scope specifying the words or phrases in the sentences that are affected by the governor. There is no correlation between the simplicity of the sentence and the number of governors. Simple sentences do not necessarily consist of a single governor, and they may have multi-governors. For example, the sentence (الشمس مشرقة., alshamsu mushriqatun, the sun is shiny) has one governor (الشمس, alshamsu, the sun) and one governed (مشرقة, mushriqatun, is shiny). The sentence (يأكل الرجل السمك, yakulu alrajulu alsamaka, the man eats the fish) has one governor (يأكل, yakulu, eats) but two governed words, (الرجل, alrajulu, the man) and (السمك, alsamaka, the fish), for the same governor. The sentence (العصفور في القفص., aleasfuru fi alqafsi, the bird is in the cage) is a nominal sentence with the predicate as an adverbial element (في القفص, fi alqafsi, in the cage). This adverbial element consists of one governor (في, fi, in) and one governed (القفص, alqafsi, the cage), but it is governed by the word (العصفور, aleasfuru, the bird), which is the (main word) in the sentence (Owens 1988). This sentence has two governors and two governed words.

3. Constraints of I`rab

This theory is based on the following set of constraints:

(a) The governed word has one and only one governor.

(b) The cardinality of the relation between the governor and the governed is one-to-many. This means that the governor has at least one governed word.

(c) The governors can be verbs, nouns or particles, but there is a set of rules used to determine the governor.

  i. All verbs are governors, regardless of their tense and type. Each verb should have a single agent that follows the verb.

  ii. Most particles are governors. Particles can be divided into three categories as follows:

    1. Particles occurring only with nouns are governors, such as the preposition particles (إلى, 'ila, to).
    2. Particles occurring only with verbs are governors, such as the particles (لم, lam, did not) and (لن, lan, will not), which occur only with the imperfect verb and govern it in the jussive and subjunctive, respectively.
    3. Particles occurring with both, such as question particles (هل, hal, whether) and conjunction particles, (و, wa, and) are not governors.

  iii. Nouns depend on its role in the sentence. For example, in the sentence (العصفور في القفص., aleasfuru fi alqafsi, the bird is in the cage), the word (العصفور, aleasfuru, the bird) is a noun, and it is a topic, so it is a governor. The word (القفص, alqafsi, the cage) is a noun, but it is an object of the prepositional (في, fi, in); therefore, it is a governed word.

4. I`rab versus dependency grammar

Mapping I`rab theory to dependency grammar is a straightforward task. An Arabic sentence can be represented as a tree of dependency relations between words. The reason behind that is that governor can be associated with the head in dependency grammar, and the governed can be associated with a modifier.

I`rab and dependency grammar have several similarities, as follows:

1. The sentence has one and only one independent word that acts as a ROOT, and the main word is a child of the ROOT node.
2. All the items are in dependency relations except the ROOT.
3. A governed word (child) has one and only one governor (parent).
4. A governor has at least one governed word.
5. A dependency relation is a unidirectional relation.

There are two major differences between I`rab and dependency grammar. The first difference concerns the determination the main word in the sentence. The dependency grammar always considers the verb as the main word, whereas the main word in I`rab depends is (mostly) the first word in the sentence. Second, the covert element(s) in the syntax representation in I`rab should be clearly deduced and depicted in the tree. In contrast, dependency grammar maps only the lexical elements in the sentence.

From the previous discussion, it can be clearly seen that although there are differences between Arabic grammatical theory and dependency grammar, I`rab has the ability to coincide with the concept of the dependency in the dependency structure.

# 3 The proposed I3rab dependency treebank

The process of building the proposed new dependency treebank passed five stages: 1) defining the mechanism of the tokenization process, 2) choosing the POS tagset, 3) describing the morphological analysis, 4) determining the I3rab dependency schema and 5) describing the format of the dependency treebank.

## 3.1 The tokenization process

The tokenization process is the task of dividing a sentence into tokens. The token is the smallest syntax unit (Attia 2007). In the Arabic language, the word is structured from concatenative morphemes that include stems, affixes and clitics. These morphemes appear sequentially in the word structure as follows (Awajan 2015)

[proclitic(s) + [prefix(es)]] + stem + [suffix(es) + [enclitic]].

The I3rab tokenization process keeps the suffix attached to the word and detaches the clitics. In general, the I3rab tokenization process detaches the following clitics from the word: question particles (أ, `a, whether), conjunction particles (و, wa, and) and (ف, fa, so/then), attached preposition particles (ب, bi, by) and future particles (س, sa, will). In addition, all joined pronouns are considered clitics and should be detached from the words. Besides that, the covert pronouns should be surmised and explicitly represented as individual tokens.

## 3.2 POS tagset

I3rab uses the same POS tagset used by PADT (Hajic et al. 2004). There are 20 tags, as listed in Table 5. The POS tags are for verbs, nouns, adjectives, adverbs, prepositions, and proper nouns. In contrast to PADT, I3rab tags all pronouns including covert pronouns and takes the (S-) POS tag.

**Table 5: The part-of-speech category and their meanings (LDC 2007)**

| Part-of-speech category | Description |
|---|---|
| VI VP VC | imperfect, perfect, and imperative verb forms |
| N- A- D- | nouns, adjectives, and adverbs |
| C- P- I- | conjunctions, prepositions, interjections |
| G- Q- Y- | graphical symbols, numbers, abbreviations |
| F- FN FI | particles, especially negative and interrogative |
| S- SD SR | pronouns, especially demonstrative and relative |
| -- | isolated definite articles |
| Z- | proper names |

## 3.3 Morphological analysis

I3rab uses the same approach of PADT, which is based on morphological information generated by MorphoTrees (Smrz and Pajas 2004) and the Lemmas and Glosses based on the Buckwalter lexicon (LDC 2004a). The most important morphological information are: mood, voice, person, gender, number, case and definiteness. The complete list of morphological information is shown in Table 6. As mentioned above, the tokens of all joined and covert pronouns should be processed by the morphological analyzer to generate their morphological features.

**Table 6: The morphological information and its meaning (LDC 2007)**

| Morphological feature | Description |
|---|---|
| Mood | Indicative, Subjunctive, or Jussive of imperfect verbs, with D if undecided between S and J |
| Voice | Active or Passive |
| Person | 1 speaker, 2 addressee, 3 others |
| Gender | morphologically overt “gender”, Masculine or Feminine |
| Number | morphologically overt “number”, Singular, Dual, or Plural |
| Case | 1 nominative, 2 genitive, 4 accusative |
| Definiteness | morphological “definiteness”, Indefinite, Definite, Reduced, or Complex |

## 3.4 I3rab dependency schema

The current I3rab dependency schema has 34 dependency relation labels listed in Table 7. All these labels are selected and derived from I‘rab theory. The process is performed by working on a set of Arabic sentences and obtaining their grammatical analysis through application of the I‘rab process by linguistic experts2. All the dependency relations are extracted from the grammatical analysis.

**Table 7: Dependency relations**

| # | Dependency Relation | Description (English) | Description (Arabic) | | # | Dependency Relation | Description (English) | Description (Arabic) | |
|---|---|---|---|---|---|---|---|---|---|
| 1 | ADJ | Adjective | صفة | | 18 | PRED-ADVP | Predicate-Adverbial phrase | خبر- شبه جملة ظرفية | |
| 2 | ADVP | Adverb | ظرف زمان/مكان | | 19 | PRED-NOUN | Predicate-Nominative noun | خبر - مفرد | |
| 3 | AGENT | Agent | فاعل | | 20 | PRED-NP | Predicate-Nominal phrase | خبر - جملة اسمية | |
| 4 | ALTER | Alternate | بدل | | 21 | PRED-PP | Predicate-Prepositional phrase | خبر شبه جملة جار ومجرور | |
| 5 | COMMA | Comma (punctuation) | ترقيم-فاصلة | | 22 | PRED-VP | Predicate-Verbal phrase | خبر - جملة فعلية | |
| 6 | COND | Condition | شرط | | 23 | PREDX-ADVP | Predicate of P-ACC or VBX-Adverbial phrase | خبر كان/إن - شبه جملة ظرفية | |
| 7 | COORD | Coordinating particle | أداة ربط | | 24 | PREDX-NOUN | Predicate of P-ACC or VBX-Nominative noun | خبر كان/إن - مفرد | |

2 Two linguistic experts were involved in this work. One has a PhD in Arabic language, and the other is in the 2nd year of a Master’s program in Arabic language.

| | | | | | | | |
|---|---|---|---|---|---|---|---|
| 8 | END | End (punctuation) | ترقيم-نقطة | 25 | PREDX-NP | Predicate of P-ACC or VBX-Nominal phrase | خبر كان/إن - جملة اسمية |
| 9 | EXCEPT | Exception | استثناء | 26 | PREDX-PP | Predicate of P-ACC or VBX-Prepositional phrase | خبر كان/إن شبه جملة جار ومجرور |
| 10 | GEN | Genitive | مجرور | 27 | PREDX-VP | Predicate of P-ACC or VBX-Verbal phrase | خبر كان/إن - جملة فعلية |
| 11 | HAAL | Adverb of manner | حال | 28 | PUNCT | Punctuation | ترقيم |
| 12 | MA3TOUF | The coordinate modifier | معطوف | 29 | TAMYEEZ | The specifier | تمييز |
| 13 | NEG (Negation) | Negation particle | حرف نفي | 30 | TAWKEED | Emphasis | توكيد |
| 14 | OBJ | Object | مفعول به (المفاعيل) | 31 | TOPIC | Topic | مبتدأ |
| 15 | P | Particle | حرف | 32 | TOPICX | Topic of P-ACC or VBX | اسم كان/إن |
| 16 | P-ACC | Accusative particle | حرف نصب | 33 | VB (Verb) | Verb (strong) | فعل تام |
| 17 | PART | Part particle | حرف تابع | 34 | VBX (copula) | Defective verb (copula) | فعل ناقص |

## 3.5 Dependency treebank format

The I3rab dependency treebank will be presented in the CoNLL-X format in a tab separated file in which sentences are separated by a blank line (Buchholz and Marsi 2006). Each sentence has one or more tokens. Each token has ten attributes separated by a tab. The ten attributes are ID (token index in the sentence), FORM (surface form of token as appears in the sentence), LEMMA (typically the lemma of the token), CPOSTAG (coarse POS), POSTAG (grain POS), FEATS (set of optional morphological features), HEAD (ID of the head of the token), DEPREL (dependency label between the HEAD and the token), PHEAD (projective head of the current token) and PDEPREL (dependency relation to the PHEAD). The columns ID, Form, HEAD and DEPREL are mandatory, and the others are optional. For example, the sentence ( وصول وزير الخارجية الامريكي الى بيروت, wusul waziri alkharijiat al'amrikii 'iilaa bayrut, US Secretary of State arrives in Beirut) is represented in CoNLL-X format, as shown in Figure 3.

| | | | | | | | | | |
|---|---|---|---|---|---|---|---|---|---|
| 1 | وُصُولُ | وُصُول_1 | N | N- | Case=1\|Defin=R | 0 | SUBJ | _ | _ |
| 2 | وَزِير | وَزِير_1 | N | N- | Case=2\|Defin=R | 1 | GEN | _ | _ |
| 3 | الخارِجِيَّةِ | خارِجِيَّة_1 | N | N- | Gender=F\|Number=S\|Case=2\|Defin=D | 2 | GEN | _ | _ |
| 4 | الأَمْرِيكِيِّ | أَمْرِيكِيّ_1 | A | A- | Case=2\|Defin=D | 2 | ADJ | _ | _ |
| 5 | إِلَى | إِلَى_1 | P | P- | _ | 1 | PRED | _ | _ |
| 6 | بَيْرُوت | بَيْرُوت_1 | Z | Z- | Case=2\|Defin=R | 5 | GEN | _ | _ |

**Figure 3: The sentence ((وصول وزير الخارجية الامريكي الى بيروت,, US Secretary of State arrives in Beirut), represented in CoNLL-X format**

The dependency labels in the example were taken from the dependency labels for the I3rab Dependency Treebank.

# 4 Implementation

To validate our approach and demonstrate the quality of the predicted grammatical structure (Solberg et al. 2014), we evaluated the I3rab treebank against the PADT treebank in a dependency parsing task. The evaluation was limited to only the PADT treebank, because the PADT (Hajic et al. 2004) treebank was involved in the CoNLL shared task 2007 (Nivre et al. 2007a; LDC 2007). In addition, the PADT treebank is available for free download from the LDC site under the LDC user agreement for non-members (LDC 2018), whereas the other MSA dependency treebanks are not available for free download.

In the validation process, we had four important components: datasets, a parser generator, accuracy metrics, and a parser evaluation tool.

1. Datasets

Two datasets were used in the evaluation throughout the experiments: the part-PADT dataset and I3rab dataset. The part-PADT dataset is a subset of PADT (Hajic et al. 2004). The PADT was mainly collected from six news agencies (Smrz, Šnaidauf, and Zemánek 2002; LDC 2004b, 2007). For the CoNLL shared task 2007, the available PADT dataset included 3043 sentences with a total of 116,800 tokens (LDC 2007). It was represented in CoNLL-X format. The original PADT dataset was divided into two sets: a training dataset including 2912 sentences and a testing dataset including 131 sentences (LDC 2018). The morphological features of tokens had been annotated by using MorphoTrees (Smrz and Pajas 2004; LDC 2007), and Lemmas and Glosses were generated according to the Buckwalter lexicon (LDC 2007, 2004a).

The part-PADT dataset contains 300 sentences. The two portions of the PADT dataset (training and testing datasets) were combined, and the sentences were selected from Xinhua News Agency (XIA).

To select the 300 sentences, we sorted the sentences of XIA according to the length of the sentence3. The sentences with lengths between 4 and 48 words were selected and were labeled with sequence IDs from 1 to 300. Below, we will use the term PADT instead of the term part-PADT. Figure 4 shows the distribution of the sentence length for the PADT.

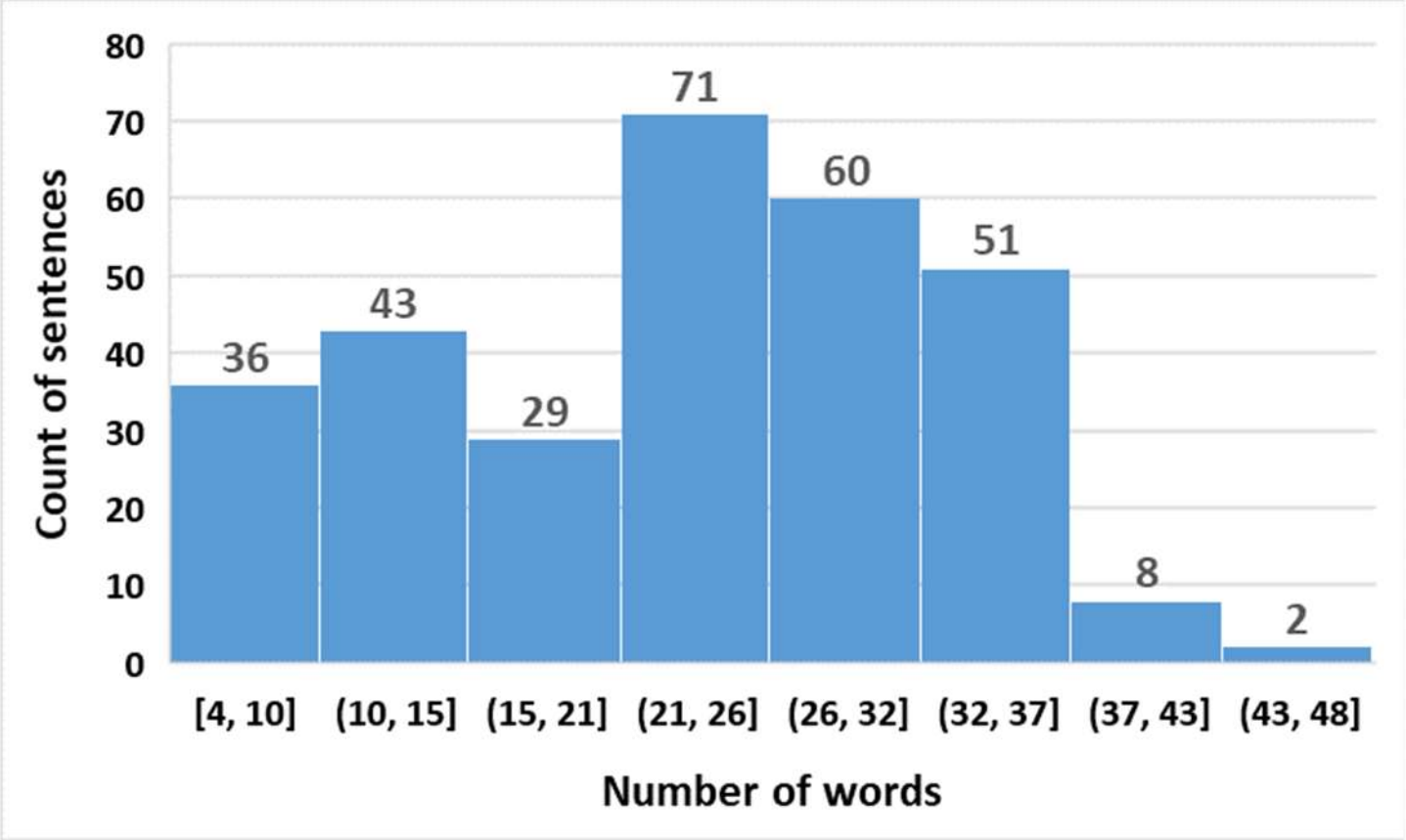

**Figure 4: The distribution of sentence length for the PADT**

The I3rab dataset was constructed by using the same 300 sentences of the PADT dataset. Initially the I3rab dataset had the same tokenization, the same POS tags and the same morphological features as the PADT dataset. The fields related to dependency relations were reinitialized to prepare them for our approach.

The tokenization of the sentences was revised and modified as required according to the tokenization approach in I3rab, especially for verbs. New tokens were generated as follows:

---

[3] The length of the sentence was measured as the number of words in the sentence; in this stage, the sentences were white-space separated.

(a) The first step was based on the theory utilized by I3rab. Each verb should have an agent, and this agent should come after the verb. This agent should be considered a separate token. According to I3rab, the agent might be a single nominative noun, independent pronoun, joined pronoun or covert pronoun.

   (i) The single nominative noun and independent pronoun: I3rab is consistent with PADT in considering them as separate tokens from the verb. For example, in the sentence (وقع الانفجار في قاع أحد المناجم حيث كان يعمل الضحايا الأربعة, waqa`a alainfijaru fi qae ahdi almanajimi haythu kana yaemalu aldahaya alarb`atu, the explosion occurred at the bottom of a mine where the four victims were working), the noun (الانفجار, alainfijaru, the explosion) is the agent of the verb (وقع, waqa`a, occurred) and is already considered a separate token according to the white-space segmentation method.

   (ii) The joined pronoun: I3rab behaves differently from the PADT in considering the joined pronoun as a separate token. For example, the word (يجتمعان, yajtamieani, they (dual) meet) was segmented into two separate tokens: the verb (يجتمع, yajtamieu, meets) and the joined nominative pronoun (ان, ani, they (dual)) that acts as the agent to the verb (يجتمع, yajtamieu, meets). Another example is the word (يدخلون, yadkhuluna, they enter), which was segmented into two separate tokens: the verb (يدخل, yadkhulu, enters) and the joined nominative pronoun (ون, una, they (plural)) that acts as an agent to the verb (يدخل, yadkhulu, enters).

   (iii) The covert pronoun: I3rab behaves differently from the PADT by allowing the agent to be surmised, despite its not being explicitly stated in the sentence. A new token was generated and added immediately after the verb. For example, in the sentence (أن لبنان يدين الإرهاب, 'anna lubnana yudinu al'iirhaba, Lebanon condemns terrorism), the verb (يدين, yudinu, condemns) has no explicit agent. The word (الإرهاب, al'iirhaba, terrorism) is a direct object of the verb, whereas the covert pronoun is considered the agent of the verb. This agent is surmised as (هو*, hiya, she), which refers to the noun (لبنان, lubnana, Lebanon).

(b) The second step was related to the tokenization process of PADT, which tokenizes some words such as (حسبما, hasbama, according to) as one token, whereas I3rab tokenizes them into two separate tokens: the word (حسب, hasba, according to) and the word (ما, ma, which). This segmentation is based on the i`rab of sentences. This separation process was performed in 29 cases.

(c) The third step was related to the existence of error in the tokenization process in the PADT. For example, the word (بالسارس, bialssarsi) should be segmented into two separate tokens: the preposition particle (ب, bi, with) and the genitive proper name (السارس, alssarsi, SARS). In another example, the word (وهواتيان, waHuatian) should be segmented into two separate tokens: the coordinating particle (و, wa, and) and the noun (هواتيان, Huatian). This process complies with the PADT philosophy. This type of error had been handled in the I3rab dataset with a frequency equal to 24. Another error is related to the definite tool in Arabic (الـ, al, the). It is well known that the PADT considers the definite tool (الـ, al, the) as a morphological feature for the noun and does not tokenize it as a separate token. There were two errors in tokenization process that occurred when the PADT isolated the definite tool from the word, especially when it was followed by a number, e.g., ((الـ, al, the) + 29). The last error was related to the existence of redundancy of some words in the text, although they did not exist in the original text. This error occurred only in one sentence. In I3rab, we deleted this word.

Table 8 shows descriptive statistics of tokens throughout the tokenization process in PADT and I3rab. The number of tokens in the I3rab dataset exceeded the number of tokens in the PADT dataset by 338, owing to the previously described tokenization process used to construct the I3rab dataset, as summarized in Table 9.

**Table 8: Descriptive statistics of tokens throughout the tokenization process in the PADT and I3rab datasets**

| --- | part-PADT | I3rab |
|---|---|---|
| Number of sentences | 300 | |
| Total number of tokens over all sentences | 6863 | 7203 |
| Minimum number of tokens per sentence | 4 | 5 |
| Maximum number of tokens per sentence | 48 | 50 |
| Average number of tokens over all sentences | 23 | 24 |
| | | |

**Table 9: The reasons behind the newly generated, merged and deleted tokens in the I3rab dataset**

| Case | Number of Cases |
|---|---|
| Dropped pronoun | 243 |
| Joined pronoun | 47 |
| Separated | 53 (29 + 24) |
| Merged | 2 (-) |
| Delete | 1 (-) |
| | 340 |

After revising the tokenization process, for each newly created or modified token, we assigned the morphological features to them on the basis of MorphoTrees. For the verbs with a joined nominative pronoun, the number feature of the verb was changed into singular, and the number feature of the pronoun was either dual or plural, depending on the joined pronoun.

After the tokenization was completed and the morphological features for the tokens were reassigned as required, the dependency relations between tokens within the sentences were constructed according to the I3rab approach. As mentioned above, there are two main differences between the I3rab approach and PADT approach. The first is determining the main word of the sentence. The second is the explicit representation of all pronoun types (independent, joined and covert).

For example, the sentence (موظفو اليونيسيف يبدأون العودة إلى بغداد, muzzafu alywnisifi yabda'uwna alawdata 'iilaa baghdada, UNICEF staff are starting to return to Baghdad) in the two datasets shows differences between PADT and I3rab. These differences can be summarized as follows: The word (يبدأون, yabda'uwna, are starting) in the PADT dataset was considered one token, and its number feature was plural. In contrast, the word (يبدأون, yabda'uwna, are starting) in the I3rab dataset was separated into two tokens: the first token was the verb (يبدأ, yabda'u, starts), and the second token was the joined nominative pronoun (ون, una, they (plural)) that acts as an agent to the verb. The number feature of the verb was singular, and the number feature of the pronoun was plural.

Moreover, according to the PADT approach, this sentence was considered a verbal sentence, and the main word of the sentence (modifier of ROOT) was the verb (يبدأون, yabda'uwna, are starting), where the ID of head was zero. In contrast, according to I3rab, this sentence was considered a nominal sentence, and because it was not preceded by any of the abolishers, the main word of the sentence was the topic (موظفو, muzzafu, staff). The CoNLL-X format of the sentence in PADT and I3rab is presented in Figures 5. The dependency trees for the sentence in PADT and I3rab are shown in Figures 6.

| PADT | | | | | | | | | |
|---|---|---|---|---|---|---|---|---|---|
| ID | FORM | LEMMA | CPOSTAG | POSTAG | FEATS | HEAD | DEPREL | PHEAD | PDEPREL |
| 1 | موظفو | موظف_1 | N | N- | Gender=M\|Number=P\|Case=1\|Defin=R | 3 | Sb | _ | _ |
| 2 | اليونيسيف | يونيسف_1 | Z | Z- | Defin=D | 1 | Atr | _ | _ |
| 3 | يبدأون | بدأ-_1 | V | VI | Mood=I\|Person=3\|Gender=M\|Number=P | 0 | Pred | _ | _ |
| 4 | العودة | عودة_1 | N | N- | Gender=F\|Number=S\|Case=4\|Defin=D | 3 | Obj | _ | _ |
| 5 | إلى | إلى_1 | P | P- | _ | 4 | AuxP | _ | _ |
| 6 | بغداد | بغداد_1 | Z | Z- | Case=2\|Defin=R | 5 | Adv | _ | _ |

**(a) In the PADT dataset**

| I3rab | | | | | | | | | |
|---|---|---|---|---|---|---|---|---|---|
| ID | FORM | LEMMA | CPOSTAG | POSTAG | FEATS | HEAD | DEPREL | PHEAD | PDEPREL |
| 1 | موظفو | موظف_1 | N | N- | Gender=M\|Number=P\|Case=1\|Defin=R | 0 | TOPIC | _ | _ |
| 2 | اليونيسيف | يونيسف_1 | Z | Z- | Defin=D | 1 | GEN | _ | _ |
| 3 | يبدأ | بدأ-_1 | V | VI | Mood=I\|Person=3\|Gender=M\|Number=S | 1 | PRED-VP | _ | _ |
| 4 | ون | هم_1 | S | S- | Person=3\|Gender=M\|Number=P\|Case=1 | 3 | AGENT | _ | _ |
| 5 | العودة | عودة_1 | N | N- | Gender=F\|Number=S\|Case=4\|Defin=D | 3 | OBJ | _ | _ |
| 6 | إلى | إلى_1 | P | P- | _ | 5 | P | _ | _ |
| 7 | بغداد | بغداد_1 | Z | Z- | Case=2\|Defin=R | 6 | GEN | _ | _ |

**(b) In the I3rab dataset**

**Figure 5: The CoNLL-X format of the sentence (موظفو اليونيسيف يبدأون العودة إلى بغداد, UNICEF staff are starting to return to Baghdad)**

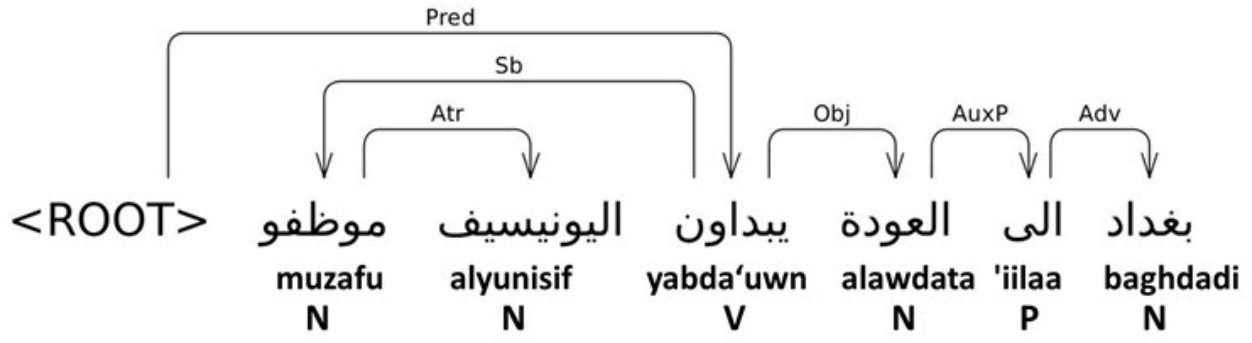

(a) In the PADT dataset

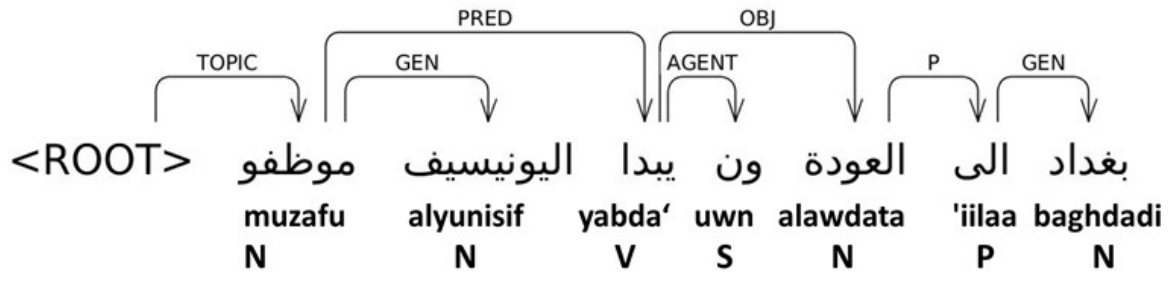

(b) In the I3rab dataset

**Figure 6: The dependency tree4 of the sentence (موظفو اليونيسيف يبدأون العودة إلى بغداد, UNICEF staff are starting to return to Baghdad)**

2. Parser generator

The MaltParser parser generator was used in our experiments to evaluate the quality of the predicted dependency structure obtained from the I3rab treebank. MaltParser is a state-of-the-art independent-language dependency parser generator. It is a shift-reduce transition based dependency parser (Nivre, Hall, and Nilsson 2006; Nivre et al. 2007b). In this paper, we used the freely available MaltParser version 1.9.2 (Nivre 2018).

3. Accuracy metrics

There is a set of metrics used to measure the quality of a dependency parser. The two most important metrics are the Unlabeled Attachment Score (UAS) and Labeled Attachment Score (LAS). UAS (head right) is the percentage of tokens that correctly linked to its head and LAS (both right) is the percentage of tokens that correctly linked to its head with right dependency relation comparing gold test data.

4. The Parser evaluation tool

In this paper, we used the free available evaluation tool MaltEval (Nilsson and Nivre 2008). It provides quantitative evaluation for the accuracy metrics (UAS, LAS) for the predicted dependency trees. The version used in this paper was released in 05/10/2014 (Johan Hall and Nivre 2013).

# 5 Experimental results and discussion

The evaluation process involved three main steps: training, testing and computation of evaluation metrics. In the first step, training, the MaltParser was elaborated (in learn mode) to produce two trained parser models: one involving the PADT training dataset and the other involving the I3rab dataset. The second step was the testing step, in which the MaltParser was elaborated (in parse mode) with the trained parser model against a blind dataset; a set of predicted grammatical structures was produced for each sentence in the testing datasets of PADT and I3rab. The blind testing dataset included the sentences for testing in CONLL-X format, with values for only six columns—ID, FORM, LEMMA, CPOSTAG, POSTAG and FEATS—for the tokens in the sentences. The dependency information (HEAD, DEPREL, PHEAD and PDEPREL) was omitted from the blind dataset. The last step was computing the evaluation metrics. In this step, MaltEval was used, and the predicted parsed trees were compared against the trees in the gold dataset for both PADT and I3rab. Then the UAS and the LAS metrics were calculated for both PADT and I3rab.

[4] In all dependency structure figures, the words are annotated with coarse POS tags. These tags are (N: noun), (V: verb), (S: pronoun), (P: particle) and (A: adjective).

In the experiments, the MaltParser training used both datasets, and testing used 10-fold cross-validation to avoid the issue of sample bias. The percentage of UAS and LAS results of 10-fold experiments for PADT and I3rab are shown in Table 10 and Table 11 respectively.

**Table 10: The percentage of UAS for 10-fold experiments for PADT and I3rab**

| Exp. # | 1 | 2 | 3 | 4 | 5 | 6 | 7 | 8 | 9 | 10 | Average |
|---|---|---|---|---|---|---|---|---|---|---|---|
| PADT | 77.4 | 78.5 | 75.4 | 75.7 | 81.8 | 78.2 | 79.2 | 76.4 | 75.2 | 80.6 | 77.8 |
| I3rab | 90.4 | 84.4 | 82.4 | 83.3 | 84.3 | 77.9 | 83.4 | 81.7 | 82.5 | 86.4 | 83.7 |

**Table 11: The percentage of LAS for 10-fold experiments for PADT and I3rab**

| Exp. # | 1 | 2 | 3 | 4 | 5 | 6 | 7 | 8 | 9 | 10 | Average |
|---|---|---|---|---|---|---|---|---|---|---|---|
| PADT | 66.8 | 65.8 | 62.0 | 63.3 | 69.4 | 66.4 | 69.7 | 65.4 | 64.4 | 70.1 | 66.3 |
| I3rab | 88.3 | 79.7 | 76.4 | 77.6 | 79.1 | 72.7 | 78.3 | 76.0 | 78.2 | 81.7 | 78.8 |

It is clearly shown that the highest UAS and LAS scores for dependency parsing for Arabic are achieved by using the I3rab dataset. The average of UAS reached 83.5, and the average of LAS reached 78.8. The percentage improvement achieved by invoking the I3rab strategy against PADT was 7.5% and 18.8% for UAS and LAS, respectively. Moreover, the differences in UAS and LAS means are considered to be extremely statistically significant ($p < 0.0005$, Paired t-test).

## 5.1 Analysis

1. Analysis - UAS

The UAS metric relates to identifying the head node of the dependent node correctly. I3rab has a higher average UAS than PADT, thus indicating that I3rab has lower syntactic complexity than PADT. The unlabeled dependency relation has two main attributes: the direction of relation and the distance[5] between head and modifier. These two attributes directly affect the value of UAS. For the first attribute, the direction of the dependency relation is the direction of the arc from the head node toward the dependent node. If the head node precedes the dependent node (Index_head < Indexdependent), then the direction of the dependency relation is RGHT. If the dependent node precedes the head node, then the direction of dependency relation is LEFT (Indexhead > Indexdependent). For example, the sentence (قبل فترة وقع حادث مماثل, qabla fatratin waqaea hadithun mumathilun, a similar incident happened a while ago) has the same dependency structure in PADT and I3rab. The token (وقع, waqaea, happened) is the head for the token (حادث, hadithun, incident), and the direction of the dependency relation is RIGHT. In contrast, the token (وقع, waqaea, happened) is simultaneously the head for the token (قبل, qabla, while ago), and the direction of the dependency relation is LEFT. The unlabeled dependency structure of the sentence is shown in Figure 7.

[5] Dependency distance = absolute ($Index_{head}$ - $Index_{dependent}$) - 1

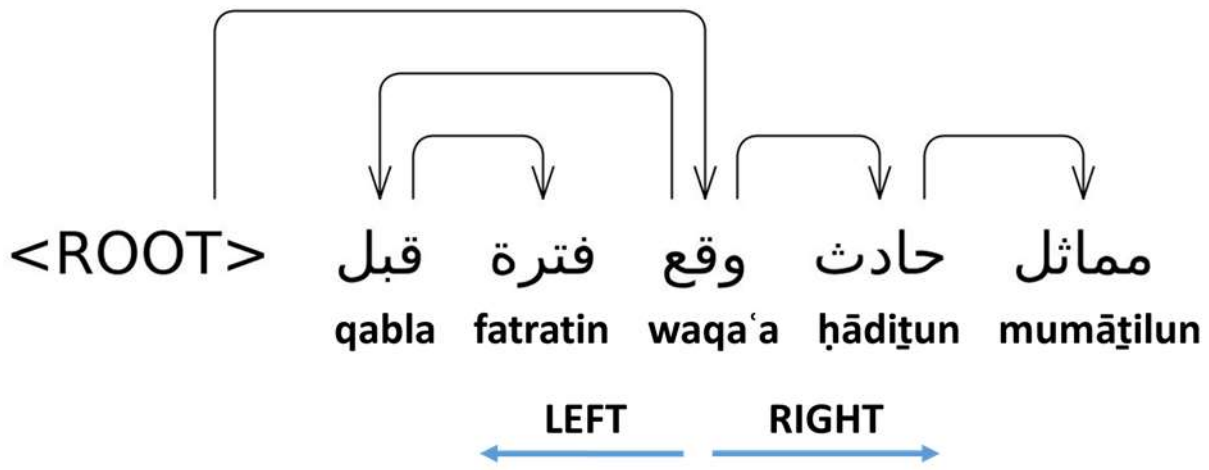

**Figure 7: The unlabeled dependency structure of the sub-sentence (قبل فترة وقع حادث مماثل, a similar incident happened a while ago)**

The percentages of both directions for the dependency relations in the PADT and I3rab datasets are shown in Table 12. Although the datasets involved were small, it is clearly shown that both datasets had a high percentage of the RIGHT direction dependency relation. In other words, the governors tend to precede the items they govern. The high percentage of both datasets showed that the Arabic language tends to support the RIGHT dependency relation more than the LEFT dependency relation.

**Table 12: The percentage of LEFT and RIGHT directions for the dependency relations in the PADT and I3rab datasets**

| | part-PADT | I3rab |
|---|---|---|
| Number of dependency relations | 6863 | 7203 |
| LEFT direction | 670 | 109 |
| % LEFT direction | 9.76% | 1.51% |
| RIGHT direction | 6193 | 7094 |
| % RIGHT direction | 90.24% | 98.49% |

I3rab had a higher percentage of RIGHT dependency relations than PADT. From the perspective of supervised data-driven dependency parser, this makes the learning and predicting the direction of dependency relation more appropriate. For the second attribute, the distance between the head and its modifiers is related to the long dependency distance problem the dependency parsing task faces. In general, if this distance is increased, then the task of linking head with its modifier(s) becomes more difficult, and the UAS value may be decreased. The long dependency distance problem can be divided into two sub-problems. The first problem is the long dependency distance between the ROOT node and main word(s) in the sentence. The second problem is the long dependency distance between the head and its modifier(s) (where the head is not the ROOT)[6]. From Figure 8 (a), I3rab shows 66% cases in which the distance between ROOT and main word is zero, whereas PADT has 50%. From Figure 8 (b), I3rab and PADT show similar distributions.

From the results we conclude that the effect of a long dependency distance between ROOT and main word(s) has a certain impact on the value of UAS. That is, as the ratio of zero distance increases, the UAS accuracy increases. This conclusion implies that the main concept of I3rab in considering the first word in the sentence as the main word will absolutely increase the ratio of zero distance between ROOT and main word(s). Moreover, because I3rab has a higher UAS ratio than PADT, then it is expected that the parser task will be easier with I3rab rather than PADT. Consequently, the annotation approach in I3rab, in general appears to decrease the complexity of the syntax structure of Arabic sentences (Nivre 2009).

[6] In order to evaluate this problem in a better way, we ignore the dependency relation between the ROOT and dot (.) at the end of sentences. This relation is ignored because both I3rab and PADT agree in considering the dot (.) at the end of the sentence as a mandatory modifier for the ROOT node, so it has the highest dependency distance compared with other tokens in the sentence, and its value is the length of the sentence.

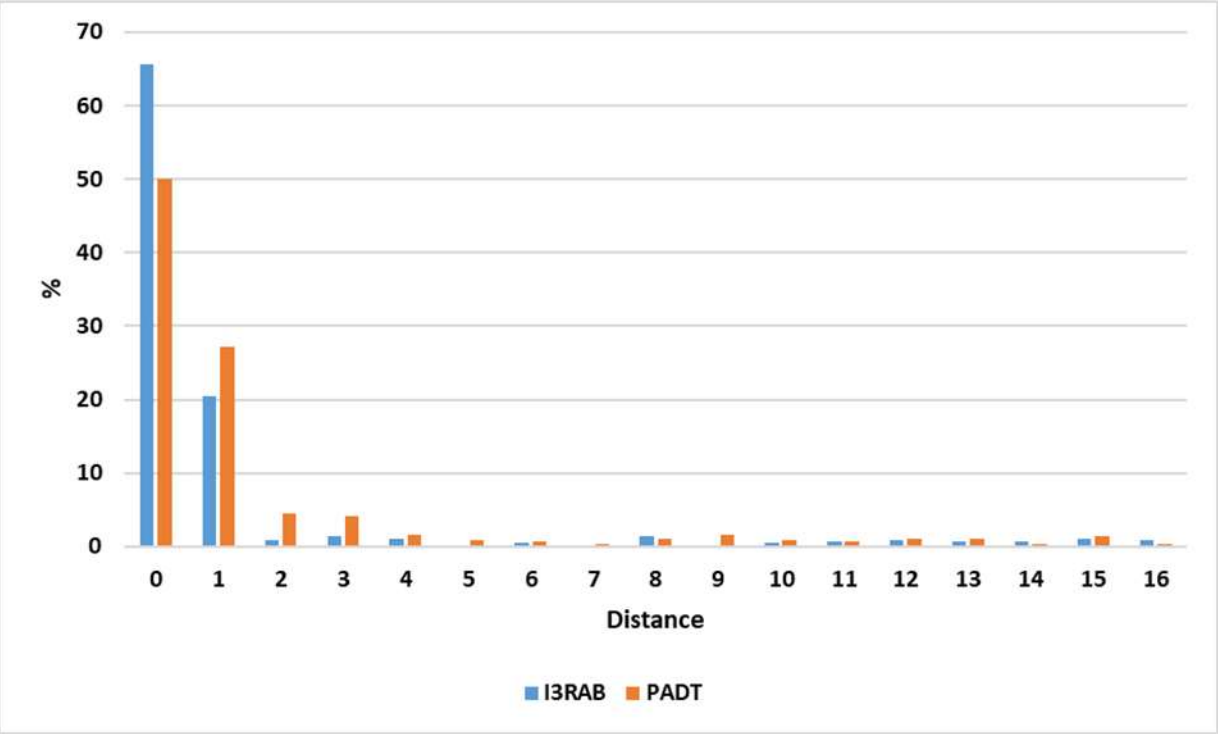

**(a) Between ROOT node and main word(s)**

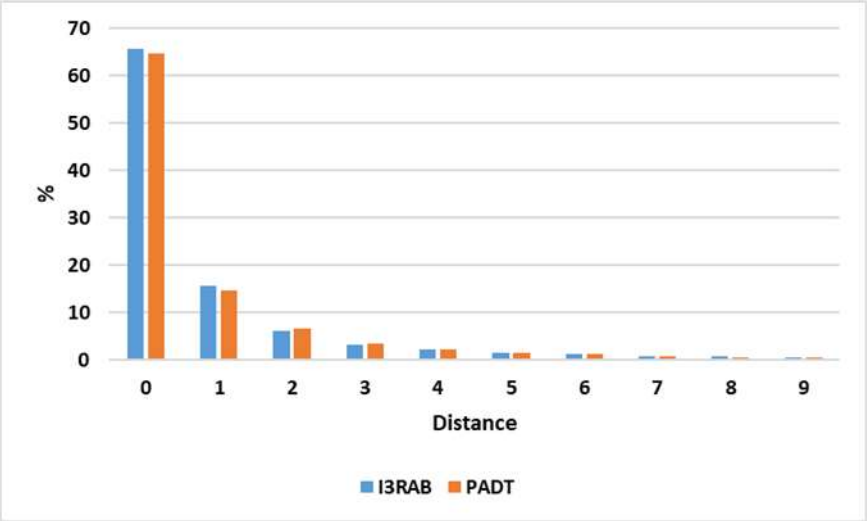

**(b) Between head and its modifier(s)**

**Figure 8: The Dependency distance distribution**

2. Analysis – LAS

The LAS metric relates to determining the type of dependency relation. In this early stage in developing I3rab treebank, we used small datasets, so many labels were sparsely represented in both datasets. The cardinality of dependency relations in both datasets was classified into five categories: very high (30-36%), high (10-15%), medium (5-9%), low (1-4%) and rare (<1%). The cardinality distribution is shown in Figure 9.

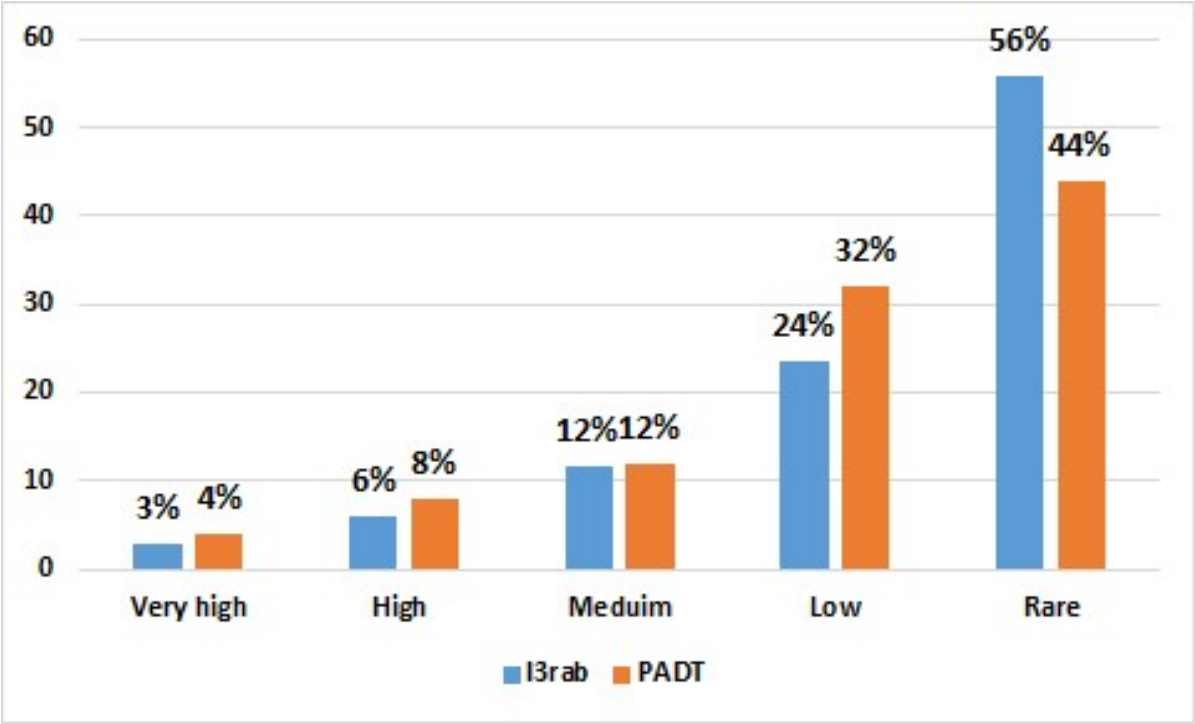

**Figure 9: The cardinality distribution of dependency relations in I3rab and PADT datasets**

In general, sparsity has a negative impact on LAS. Although, the distribution in I3RAB was worse than the distribution in PADT, but I3rab has a higher average LAS than PADT. This higher value mainly related to the higher value of average UAS. We expect to achieve an improvement in LAS as the size of datasets is increased. However, in the case of adding longer and more complex sentences, the UAS is expected to be dropped.

## 5.2 Discussion

This paper argues that the main reasons behind the higher value of UAS and LAS for I3rab than PADT are related to the concepts of determining the main word of the sentence and the explicit presentation for all pronouns: independent, joined or covert. In this section, we discuss the common linguistic structure of Arabic sentences:

1. The nominal sentence
    a. The pure nominal sentence

The pure nominal sentence has no verb at all. It is a common and frequent linguistic structure in Arabic sentences. In the case of a pure nominal sentence that is not introduced by an abolisher, I3rab and PADT use a similar approach in constructing the dependency structure. For example, in the sentence (وصول وزير الخارجية الامريكي الى بيروت, wusulu waziri alkharijiati al'amrikii 'iilaa bayruta, US Secretary of State arrives in Beirut), both PADT and I3rab consider the topic (وصول, wusulu, arrives) as the main word in the sentence and link it with ROOT. They also link the predicate as a modifier to the topic[7]. The PADT and I3rab dependency structures of the sentence (وصول وزير الخارجية الامريكي الى بيروت, US Secretary of State arrives in Beirut) are shown in Figure 10.

[7] It is worth mentioning that in the original paper of PADT, PADT follow the approach of considering the predicate (\<إلى>) as the main word in the sentence, because they argued that the predicate is not usually omitted in the sentence, but the topic could be omitted in some cases. This sentence is extracted from the PADT freely available from LDC.

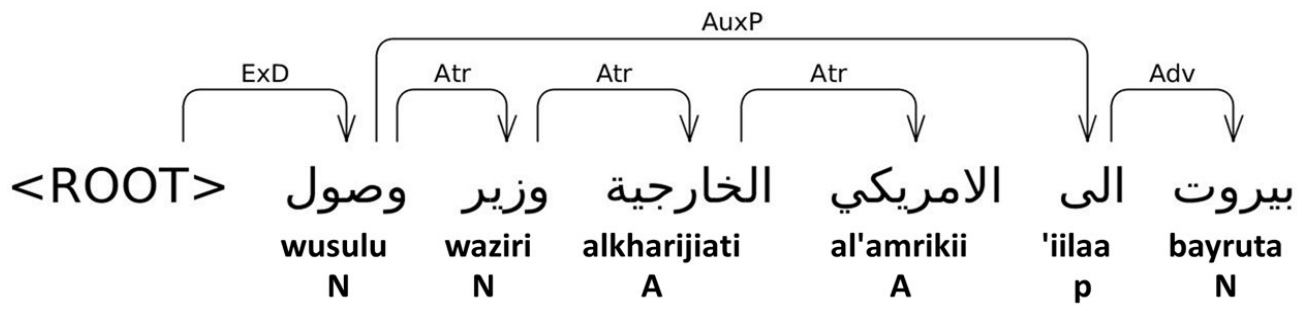

**(a) In the PADT dataset**

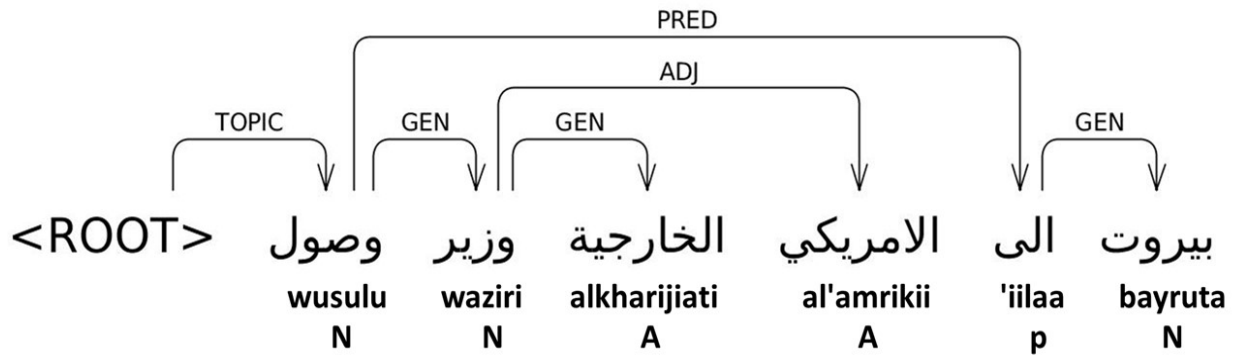

**(b) In the I3rab dataset**

**Figure 10: The dependency tree for the sentence (وصول وزير الخارجية الامريكى الى بيروت, US Secretary of State arrives in Beirut)**

In this case of pure nominal sentence I3rab and PADT follow the approach of considering the topic of the sentence as the main word in the sentence. This approach reduces the distance between the ROOT and its modifiers that implies to avoid the long distance between the ROOT and its modifiers. It is worth to mention that in nominal sentences, mostly the topic is the first word in the sentence. It often becomes before the predicate although there are some cases that the predicate become before the topic.

In the case of a pure nominal sentence introduced by an abolisher of the Inna-its-sister type, PADT and I3rab do not use the same approach in constructing the dependency structure. In the sentence ( ان العراقيين قادرون على تقرير مصيرهم بأنفسهم, 'inna aleiraqiiyna qadiruna alaa taqriri masiri him bianfshim, Iraqis are capable of self-determination), the PADT approach links the predicate (قادرون, qadiruna, are capable) as a modifier to the abolisher (ان, 'inna, that) and links the topic (العراقيين, aleiraqiiyna, Iraqis) as a modifier to the predicate (قادرون, qadiruna, are capable). In contrast, the I3rab approach links the topic (العراقيين, aleiraqiiyna, Iraqis) and the predicate (قادرون, qadiruna, are capable) to the abolisher (ان, 'inna, that) as its modifiers. The dependency structure of the sentence is illustrated in Figure 11. By comparing PADT with I3rab, we find that PADT increases the dependency distance, whereas I3rab has minimal dependency distance. Moreover, PADT adopts the LEFT direction for the dependency relation, but I3rab keeps the RIGHT direction for the dependency relation.

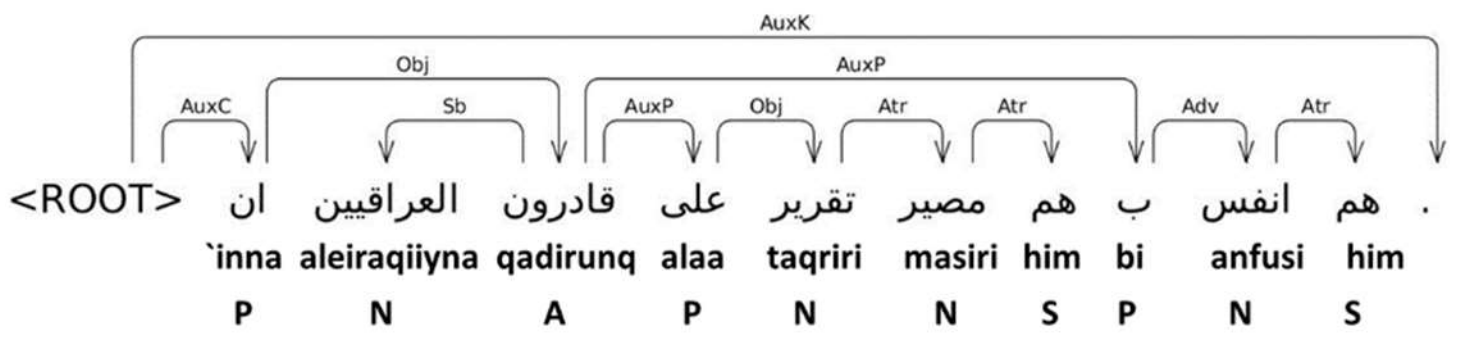

**(a) In the PADT dataset**

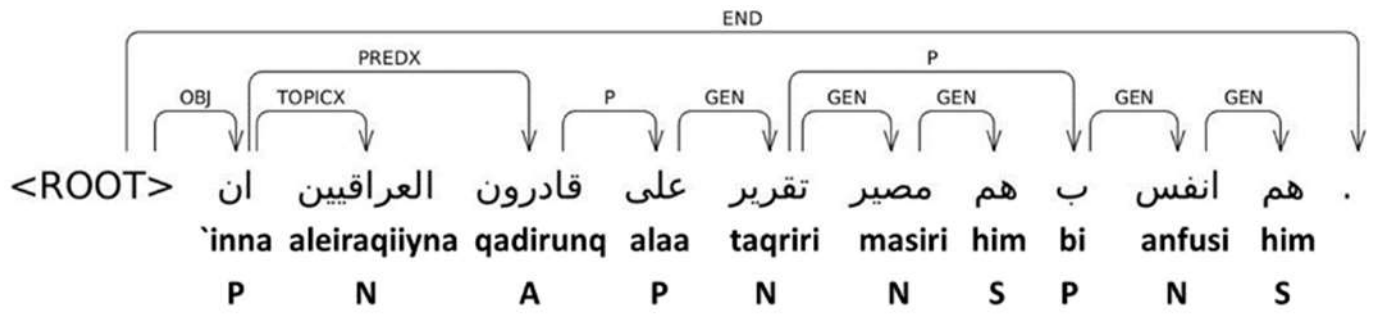

**(b) In the I3rab dataset**

**Figure 11: The dependency tree for the sentence (ان العراقيين قادرون على تقرير مصيرهم بأنفسهم., Iraqis are capable of self-determination)**

For abolishers of the Kana-its-sister type, PADT and I3rab use a similar approach in constructing the dependency structure. As shown in the sentence (كان محمد يقرأ الكتاب الجديد, kana muhamadun yaqrau alkitaba aljadida, Mohammad was reading the new book), the dependency structure of the sentence in both I3rab and PADT links the topic (محمد, muhamadun, Mohammed) and the predicate (يقرأ, yaqrau, reads) to abolisher (كان, kana, was) as modifiers. In this case, both PADT and I3rab have minimal dependency distance and keep the RIGHT direction for the dependency relation. The PADT and I3rab dependency structure of the sentence is illustrated in Figure 12.

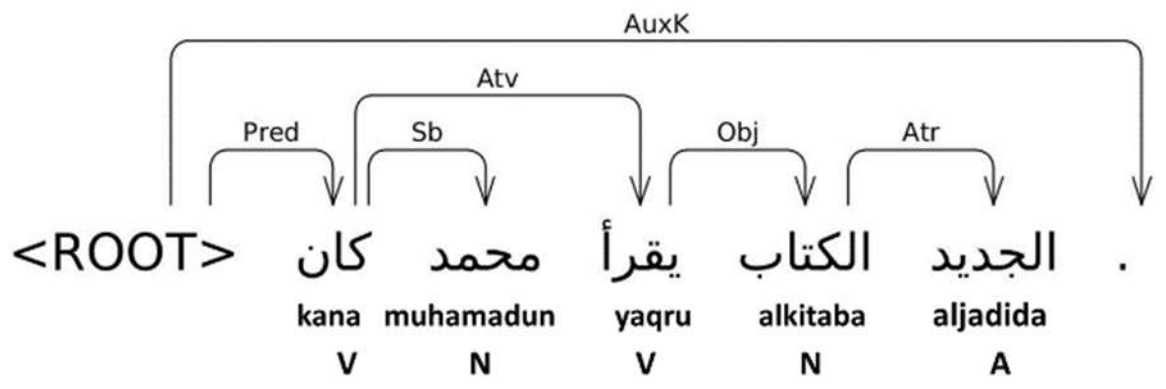

**(a) In the PADT dataset**

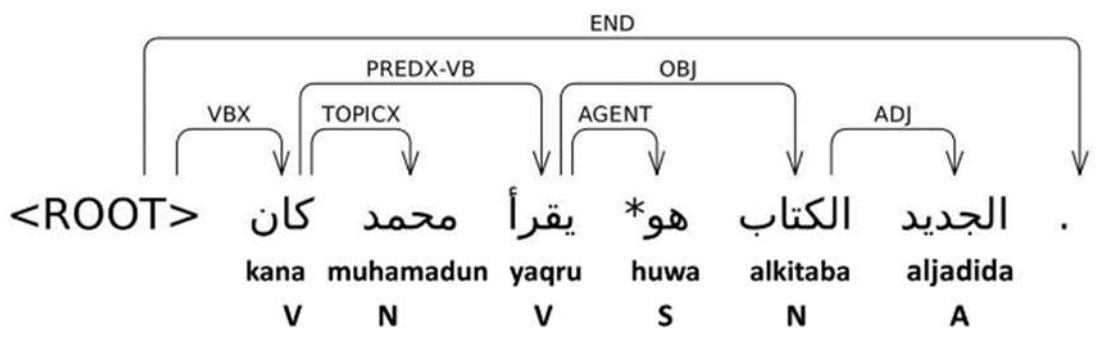

**(b) In the I3rab dataset**

**Figure 12: The dependency tree for the sentence (كان محمد يقرأ الكتاب الجديد., Mohammad was reading the new book)**

b. The nominal sentence with a predicate as a verbal sentence

Constructing the dependency structure of this type of sentence is one of the major differences between the PADT and I3rab approaches. For example, in the sentence (وزير الخارجية الامريكى يرحب بمبادرة السلام الهندية, waziru alkharijiati al'amrikii yurahibu bimubadarati alsalami alhindiati, US Secretary of State Welcomes Indian peace initiative), PADT considers the verb (يرحب, yurahibu, welcomes) as the main word in the sentence and associates it with the ROOT as modifier. In addition, it considers the word (وزير, waziru, minister) as a subject that precedes the verb and links it to the verb (يرحب, yurahibu, welcomes) as a modifier with the LEFT direction for the dependency relation. However, I3rab considers this sentence as a nominal sentence that is not introduced by any of abolishers. I3rab considers the topic (وزير, waziru, minister) as the main word in the sentence and links it with the ROOT node as a modifier. I3rab also addresses the concept that each verb should have a subject, and this subject must come after the verb. In this sentence, the verb (يرحب, yurahibu, welcomes) has a covert subject pronoun that is surmised to be (هو*, huwa, he) and is explicitly represented as an individual token in the sentence. This covert pronoun is linked to the verb (يرحب, yurahibu, welcomes) as an agent with the RIGHT direction for the dependency relation. The dependency structure of the sentence following PADT and I3rab approaches is shown in Figure 13.

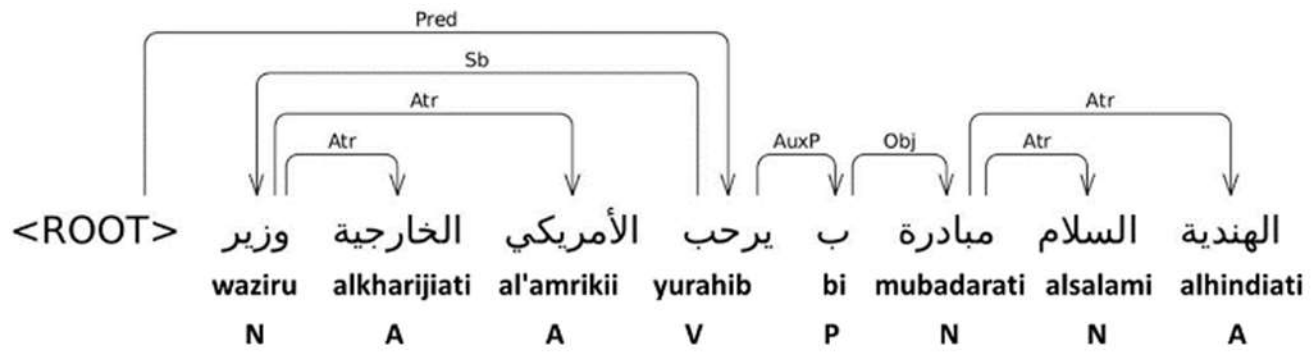

**(a) In the PADT dataset**

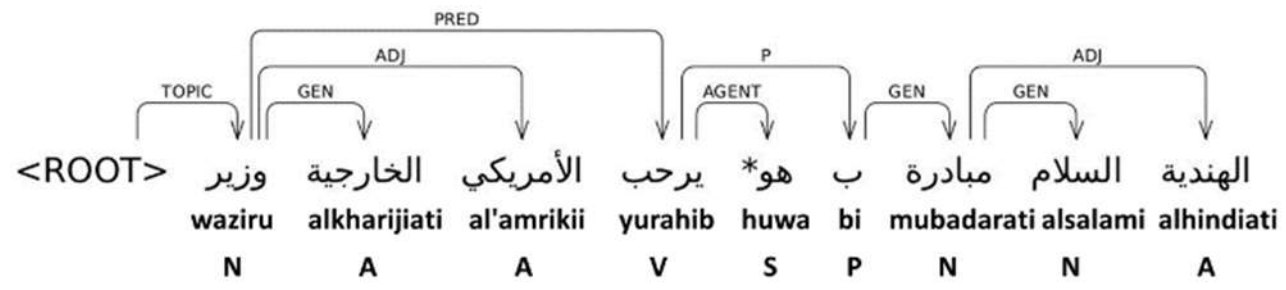

**(b) In the I3rab dataset**

**Figure 13: The dependency tree for the sentence (وزير الخارجية الامريكى يرحب بمبادرة السلام الهندية, US Secretary of State welcomes Indian peace initiative)**

By comparing PADT with I3rab in the case of sentences with a verb, we find that PADT increases the dependency distance while I3rab has minimal dependency distance. Moreover, PADT adopts the LEFT direction for the dependency relation, but I3rab keeps the RIGHT direction for the dependency relation.

2. The verbal sentence

In a sentence starting with a verb, both PADT and I3rab use a similar approach. For example, in the sentence (يلتحق منتسبو الشرطة المحلية ببغداد بمقرات عملهم السابقة, yaltahiqu muntasibu alshurtati almahaliyati bi baghdada bi makatibi amalihim alssabiqati, Local police officers join Baghdad at their former headquarters), the subject (منتسبو, muntasibu, officers) is linked as a modifier to the verb (يلتحق, yaltahiqu, join). The dependency structure of the sentence following the PADT and I3rab approaches is shown in Figure 14. In this case, both PADT and I3rab have minimal dependency distance and keep the RIGHT direction for the dependency relation.

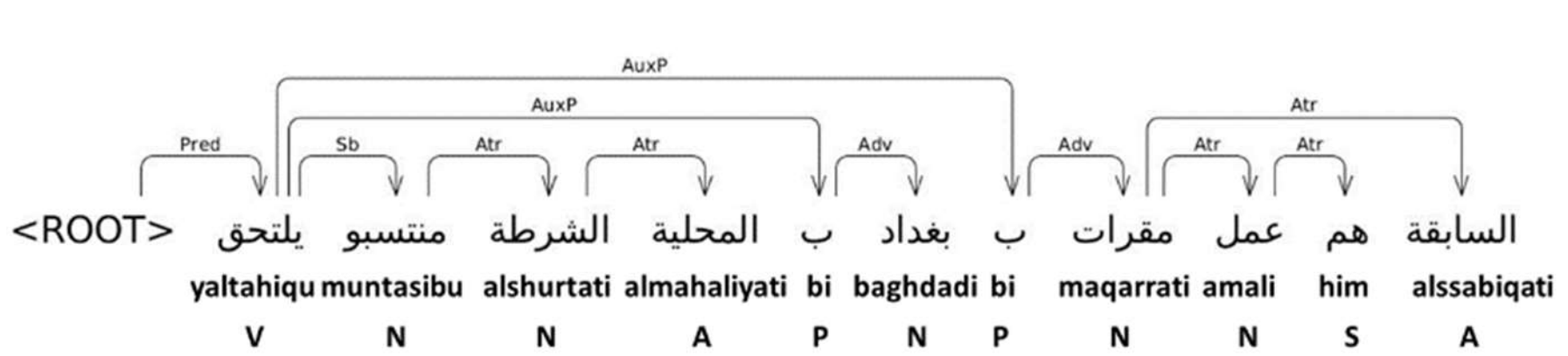

(a) In the PADT dataset

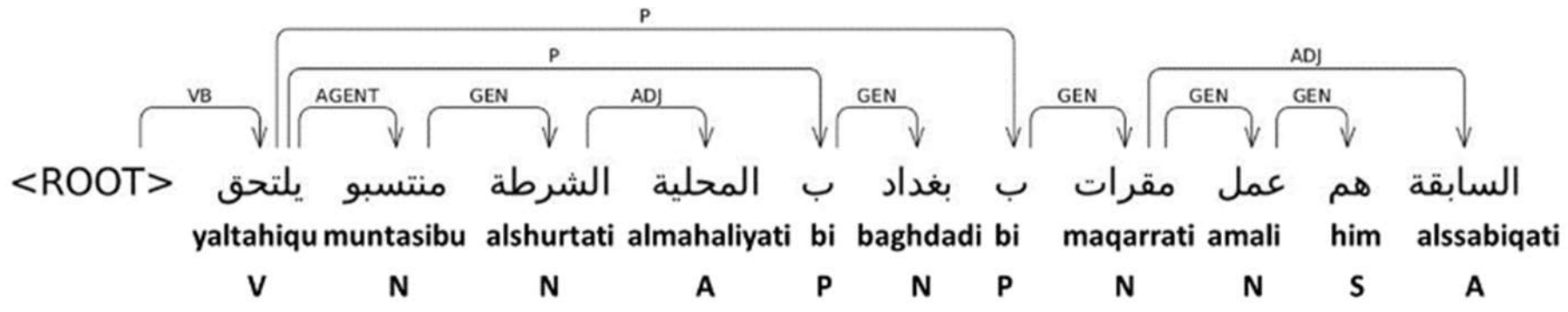

**(b) In the I3rab dataset**

**Figure 14: The dependency tree for the sentence (يلتحق منتسبو الشرطة المحلية ببغداد بمقرات عملهم السابقة., Local police officers join Baghdad at their former headquarters)**

However, if the sentence starts with an accusative or a jussive particle, the behavior in PADT and I3rab differs. For example, in the sentence (لن يقرأ محمد الكتاب ليلاً, lan yaqra`a muhammadun alkitababa laylan, Muhammad will not read the book at night), PADT considers the accusative particle (لن, lan, will not) as a modifier of the verb (يقرأ, yaqra`a, reads), and the direction of dependency relation is LEFT. I3rab considers the accusative particle (لن, lan, will not) as a head to the verb (يقرأ, yaqra`a, reads), and the direction of dependency relation is RIGHT. The dependency structure of the sentence following PADT and I3rab approaches is shown in Figure 15.

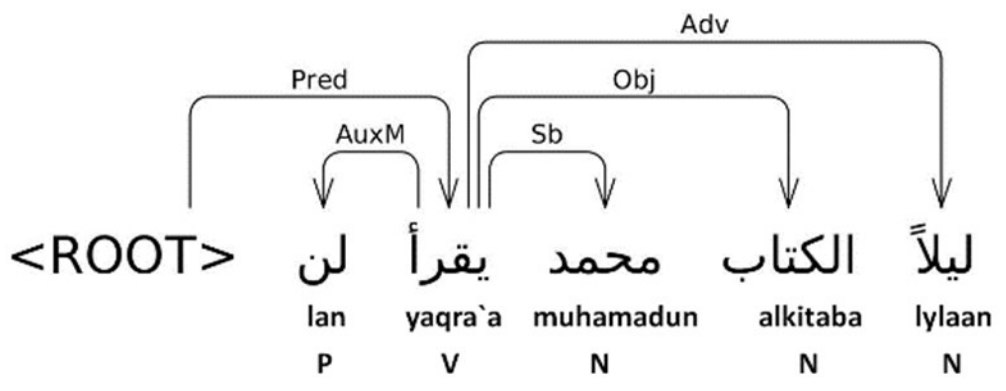

**(a) In the PADT dataset**

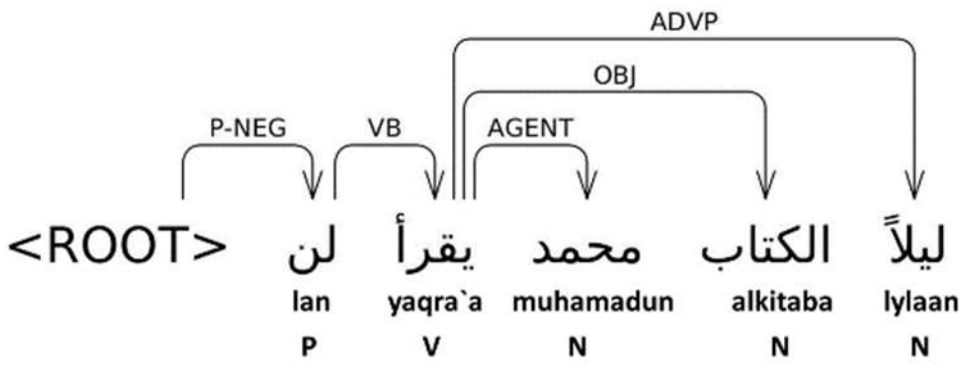

**(b) In the I3rab dataset**

**Figure 15: The dependency tree for the sentence (لن يقرأ محمد الكتاب ليلاً, Muhammad will not read the book at night)**

The same issue is repeated with a jussive particle. In the sentence (لم يقرأ محمد الكتاب في المكتبة, lam yaqra` muhammadun alkitaba fi almaktabati, Muhammad did not read the book the library), PADT considers the jussive particle (لم, lam, did not) as a modifier of the verb (يقرأ, yaqra`, reads), and the direction of the dependency relation is LEFT. I3rab considers the jussive particle (لم, lam, did not) as the head to verb, and the direction of dependency relation is RIGHT. The dependency structure of the sentence following PADT and I3rab approaches is shown in Figure 16.

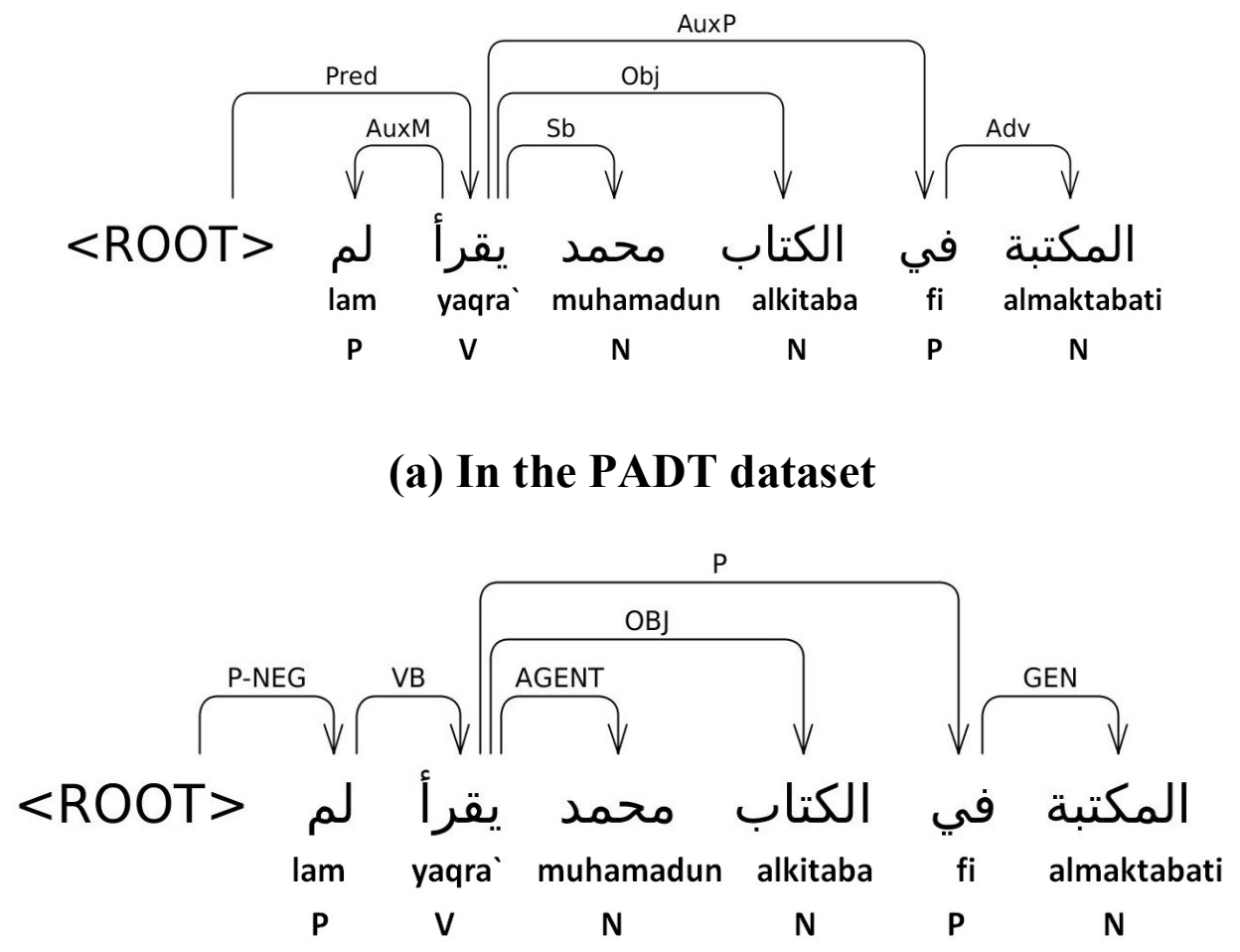

**(a) In the PADT dataset**

**(b) In the I3rab dataset**

**Figure 16: The dependency tree for the sentence (لم يقرأ محمد الكتاب في المكتبة, Muhammad did not read the book the library)**

By comparing PADT with I3rab in the case of verbal sentences that start with an accusative or a jussive particle, PADT increases the dependency distance, whereas I3rab has minimal dependency distance. In addition, PADT adopts the LEFT direction for the dependency relation, but I3rab keeps the RIGHT direction for the dependency relation.

# 6 Conclusion and future work

In this article, we have presented a new dependency treebank for Arabic language that allows the syntactic tree to reflect the real characteristics of Arabic sentences. The performance of the dependency parser using the newly constructed I3rab treebank was compared with its performance using the PADT which is the competitive dependency treebank for Arabic. We have demonstrated the effectiveness of our approach of constructing dependency treebank based on the concepts and theories identifying the linguistic structure of Arabic. In general, the I3rab approach tends to provide minimal dependency distance and to keep the direction of dependency relation to the RIGHT direction as much as possible. The minimal dependency distance simplifies the parser task, because dealing with small distances is easier than dealing with long distanced. However, keeping one direction for the most dependency relations increases the stability in training the parser model. The results showed that we gained an improvement of 7.5% and 18.8% in UAS and LAS, respectively, by using I3rab rather than PADT.

In future work, we plan to perform Inter-annotator agreement to improve the current annotation guidelines in addition to enlarging the size of the treebank. Increasing the size of the treebank will include longer statements and linguistic structures that are rarely seen in the current treebank. We are also interested in developing the universal version of the I3rab dependency treebank to facilitate cross-lingual studies, and to compare the new Arabic dependency treebank with the existing resources for Arabic and other languages.

## Appendix A: Nominative personal pronouns (independent, joined, covert)

| Person | first | | second | | | | | third | | | | | |
|---|---|---|---|---|---|---|---|---|---|---|---|---|---|
| Number | Singular | Dual/Plural | Singular | | Dual | Plural | | Singular | | Dual | | Plural | |
| Gender | Masculine/ Feminine | Masculine/ Feminine | Masculine | Feminine | Masculine/ Feminine | Masculine | Feminine | Masculine | Feminine | Masculine | Feminine | Masculine | Feminine |
| Independent Pronoun | أنا | نحن | أنتَ | أنتِ | أنتما | أنتم | أنتن | هو | هي | هما | هما | هم | هن |
| Perfect | كتبتُ | كتبنا | كتبتَ | كتبتِ | كتبتما | كتبتم | كتبتُنَّ | كتبَ | كتبتْ | كتبا | كتبتا | كتبوا | كتبنَ |
| Joined Pronoun | تُ (تاء الفاعل) | نا | تَ (تاء الفاعل) | تِ (تاء الفاعل) | تُ (تاء الفاعل) | تُ (تاء الفاعل) | تُ (تاء الفاعل) | ---- | ---- | ا (ألف الإثنين) | ا (ألف الإثنين) | و (واو الجمع) | ن (نون النسوة) |
| Example with Perfect | أنا كتبتُ المقالةَ. | نحن كتبنا المقالةَ. | أنتَ كتبتَ المقالةَ. | أنتِ كتبتِ المقالةَ. | أنتما كتبتما المقالةَ. | أنتم كتبتم المقالةَ. | أنتن كتبتُنَّ المقالةَ. | هو كتبَ المقالةَ. | هي كتبتْ المقالةَ. | هما كتبا المقالةَ. | هما كتبتا المقالةَ. | هم كتبوا المقالةَ. | هن كتبنَ المقالةَ. |
| Term segmentation | كتبتُ = كتب + تُ | كتبنا = كتب + نا | كتبتَ = كتب + تَ | كتبتِ = كتب + تِ | كتبتما = كتب + تُما** | كتبتم = كتب + تُم** | كتبتُنَّ = كتب + تُنَّ** | ---- | ---- | كتبا = كتب + ا | كتبتا = كتب +ت + ا | كتبوا = كتب + وا***** | كتبنَ = كتب + ن |
| Covert Pronoun | ---- | ---- | ---- | ---- | ---- | ---- | ---- | هو* | هي* | ---- | ---- | ---- | ---- |
| Example with Perfect (show covert) | ---- | ---- | ---- | ---- | ---- | ---- | ---- | هو كتبَ (هو*) المقالةَ. | هي كتبتْ (هي*) المقالةَ. | ---- | ---- | ---- | ---- |
| Imperfect | أكتبُ | نكتبُ | تكتبُ | تكتبينَ | تكتبان | تكتبونَ | تكتبنَ | يكتبُ | تكتبُ | يكتبان | تكتبان | يكتبونَ | يكتبنَ |
| Joined Pronoun | ---- | ---- | ---- | ي (ياء المخاطبة) | ا (ألف الإثنين) | و (واو الجمع) | ن (نون النسوة) | ---- | ---- | ا (ألف الإثنين) | ا (ألف الإثنين) | و (واو الجمع) | ن (نون النسوة) |
| Example with Imperfect | أنا أكتبُ المقالةَ. | نحن نكتبُ المقالةَ. | أنتَ تكتبُ المقالةَ. | أنتِ تكتبينَ المقالةَ. | أنتما تكتبانِ المقالةَ. | أنتم تكتبونَ المقالةَ. | أنتن تكتبنَ المقالةَ. | هو يكتبُ المقالةَ. | هي تكتبُ المقالةَ. | هما يكتبان المقالةَ. | هما تكتبانِ المقالةَ. | هم يكتبونَ المقالةَ. | هن يكتبنَ المقالةَ. |
| Term segmentation | ---- | ---- | ---- | تكتبين = تكتب + ين | تكتبان = تكتب + ان*** | تكتبون = تكتب + ون**** | تكتبن = تكتب + ن | ---- | ---- | يكتبان = يكتب + ان | تكتبان = تكتب + ان*** | يكتبون = يكتب + ون**** | يكتبن = يكتب + ن |
| Covert Pronoun | أنا* | نحن* | أنتَ* | ---- | ---- | ---- | ---- | هو* | هي* | ---- | ---- | ---- | ---- |
| Example with Imperfect (show covert) | أنا أكتبُ (أنا*) المقالةَ. | نحن نكتبُ (نحن*) المقالةَ. | أنتَ تكتبُ (أنتَ*) المقالةَ. | ---- | ---- | ---- | ---- | هو يكتبُ (هو*) المقالةَ. | هي تكتبُ (هي*) المقالةَ. | ---- | ---- | ---- | ---- |
| Imperative | ---- | ---- | اكتبْ | اكتبي | اكتبا | اكتبوا | اكتبنَ | ---- | ---- | ---- | ---- | ---- | ---- |
| Joined Pronoun | ---- | ---- | ---- | ي (ياء المخاطبة) | ا (ألف الإثنين) | و (واو الجمع) | ن (نون النسوة) | ---- | ---- | ---- | ---- | ---- | ---- |
| Example with Imperative | ---- | ---- | أنتَ اكتبْ المقالةَ. | أنتِ اكتبي المقالةَ. | أنتما اكتبا المقالةَ. | أنتم اكتبوا المقالةَ. | أنتن اكتبنَ المقالةَ. | ---- | ---- | ---- | ---- | ---- | ---- |
| Term segmentation | ---- | ---- | ---- | اكتبي = اكتب + ي | اكتبا = اكتب + ا | اكتبوا = اكتب + وا***** | اكتبنَ = اكتب + ن | ---- | ---- | ---- | ---- | ---- | ---- |
| Covert Pronoun | ---- | ---- | أنتَ* | ---- | ---- | ---- | ---- | ---- | ---- | ---- | ---- | ---- | ---- |
| Example with Imperative (show covert) | ---- | ---- | أنتَ اكتبْ (أنتَ*) المقالةَ. | ---- | ---- | ---- | ---- | ---- | ---- | ---- | ---- | ---- | ---- |

**Figure 1: Nominative personal pronouns (independent, joined, covert)**

## Appendix B: Accusative and genitive personal pronouns (independent, joined, covert)

| Person | first | | second | | | | | third | | | | |
|---|---|---|---|---|---|---|---|---|---|---|---|---|
| Number | Plural | Singular | Singular | | Dual | Plural | | Singular | | Dual | Plural | |
| Gender | Masculine/ Feminine | Masculine/ Feminine | Masculine | Feminine | Masculine/ Feminine | Masculine | Feminine | Masculine | Feminine | Masculine/ Feminine | Masculine | Feminine |
| Joined Pronoun | نا | ي (ياء المتكلم) | ك (كاف الخطاب) | | | | | ه (هاء الغائب) | | | | |
| | | | كَ | كِ | كُ | كُ | كُ | هُ | هَ | هُ | هُ | هُ |
| Example with verb | المعلم يسألنا. | المعلم يسألني. | المعلم يسألكَ. | المعلم يسألكِ. | المعلم يسألكما. | المعلم يسألكم. | المعلم يسألكن. | المعلم يسأله. | المعلم يسألها. | المعلم يسألهما. | المعلم يسألهم. | المعلم يسألهن. |
| Term segmentation | يسألنا = يسأل + نا | يسألني = يسأل + ني* | يسألكَ = يسأل + كَ | يسألكِ = يسأل + كِ | يسألكما = يسأل + كما** | يسألكم = يسأل + كم** | يسألكن = يسأل + كن** | يسأله = يسأل + هُ | يسألها = يسأل + ها*** | يسألهما = يسأل + هما*** | يسألهم = يسأل + هم*** | يسألهن = يسأل + هن*** |
| Example with noun | قرأ المعلم مقالتنا. | قرأ المعلم مقالتي. | قرأ المعلم مقالتكَ. | قرأ المعلم مقالتكِ. | قرأ المعلم مقالتكما. | قرأ المعلم مقالتكم. | قرأ المعلم مقالتكن. | قرأ المعلم مقالتهُ. | قرأ المعلم مقالتها. | قرأ المعلم مقالتهما. | قرأ المعلم مقالتهم. | قرأ المعلم مقالتهن. |
| Term segmentation | مقالتنا = مقالة + نا | مقالتي = مقالة + ي* | مقالتكَ = مقالة + كَ | مقالتكِ = مقالة + كِ | مقالتكما = مقالة + كما** | مقالتكم = مقالة + كم** | مقالتكن = مقالة + كن** | مقالتهُ = مقالة + هُ | مقالتها = مقالة + ها*** | مقالتهما = مقالة + هما*** | مقالتهم = مقالة + هم*** | مقالتهن = مقالة + هن*** |

**Figure 1: Accusative and genitive personal pronouns (independent, joined, covert)**

## REFERENCES

[1] Al-Sughaiyer, Imad A and Ibrahim A Al-Kharashi. 2004. Arabic morphological analysis techniques: A comprehensive survey. JOURNAL OF THE AMERICAN SOCIETY FOR INFORMATION SCIENCE AND TECHNOLOGY, 55(3):189–213.

[2] Alosh, Mahdi. 2005. Using Arabic: A guide to contemporary usage. Cambridge University Press Cambridge.

[3] Alotaiby, Fahad, Salah Foda, and Ibrahim Alkharashi. 2010. Clitics in Arabic language: a statistical study. In Proceedings of the 24th Pacific Asia Conference on Language, Information and Computation.

[4] Ambati, Vamshi. 2008. Dependency structure trees in syntax based machine translation. In Adv. MT Seminar Course Report, volume 137.

[5] Attia, Mohammed and Harold Somers. 2008. Handling Arabic morphological and syntactic ambiguity within the LFG framework with a view to machine translation, volume 279. University of Manchester Manchester.

[6] Attia, Mohammed A. 2007. Arabic tokenization system. In Proceedings of the 2007 workshop on computational approaches to Semitic languages: Common issues and resources, pages 65–72, Association for Computational Linguistics.

[7] Awajan, Arafat. 2007. Arabic text preprocessing for the natural language processing applications. Arab Gulf Journal of Scientific Research, 25(4):179–189.

[8] Awajan, Arafat. 2015. Keyword extraction from Arabic documents using term equivalence classes. ACM Transactions on Asian and Low-Resource Language Information Processing, 14(2):7.

[9] Awajan, Arafat. 2016. Semantic similarity based approach for reducing Arabic texts dimensionality. International Journal of Speech Technology, 19(2):191–201.

[10] Bengoetxea, Kepa and Koldo Gojenola. 2010. Application of different techniques to dependency parsing of basque. In Proceedings of the NAACL HLT 2010 First Workshop on Statistical Parsing of Morphologically-Rich Languages, pages 31–39, Association for Computational Linguistics.

[11] Bharati, Akshar, Vineet Chaitanya, Rajeev Sangal, and KV Ramakrishnamacharyulu. 1995. Natural language processing: a Paninian perspective. Prentice-Hall of India New Delhi.

[12] Bikel, Daniel M. 2004. On the parameter space of generative lexicalized statistical parsing models. Citeseer. Böhmová, Alena, Jan Hajiˇc, Eva Hajiˇcová, and Barbora Hladká. 2003. The Prague dependency treebank. In Treebanks. Springer, pages 103–127.

[13] Bouma, Gosse, Jori Mur, Gertjan Van Noord, Lonneke Van Der Plas, and Jörg Tiedemann. 2005. Question answering for dutch using dependency relations. In Workshop of the Cross-Language Evaluation Forum for European Languages, pages 370–379, Springer.

[14] Buchholz, Sabine and Erwin Marsi. 2006. Conll-x shared task on multilingual dependency parsing. In Proceedings of the tenth conference on computational natural language learning, pages 149–164, Association for Computational Linguistics.

[15] Chomsky, Noam. 1993. Lectures on government and binding: The Pisa lectures. 9. Walter de Gruyter.

[16] Chomsky, Noam and DavidW Lightfoot. 2002. Syntactic structures. Walter de Gruyter.

[17] Civit, M, N Bufí, and P Valverde. 2004. Cat3lb: a treebank for Catalan with word sense annotation. In 3rd Workshop on Treebanks and Linguistic Theories. Tuebingen, Germany.

[18] Civit, Montserrat and Ma Antònia Martí. 2004. Building cast3lb: A Spanish treebank. Research on Language and Computation, 2(4):549–574.

[19] Collins, Michael. 2003. Head-driven statistical models for natural language parsing. Computational linguistics, 29(4):589–637.

[20] Comas, Pere R, Jordi Turmo, and Lluís Márquez. 2010. Using dependency parsing and machine learning for factoid question answering on spoken documents. In Eleventh Annual Conference of the International Speech Communication Association.

[21] Dukes, Kais. 2015. Statistical parsing by machine learning from a classical Arabic treebank. arXiv preprint arXiv:1510.07193.

[22] Dukes, Kais and Tim Buckwalter. 2010. A dependency treebank of the Quran using traditional Arabic grammar. In 2010 the 7th International Conference on Informatics and Systems (INFOS), pages 1–7, IEEE.

[23] Dukes, Kais and Nizar Habash. 2011. One-step statistical parsing of hybrid dependency-constituency syntactic representations. In Proceedings of the 12th International Conference on Parsing Technologies, pages 92–103, Association for Computational Linguistics.

[24] Farghaly, Ali and Khaled Shaalan. 2009. Arabic natural language processing: Challenges and solutions. ACM Transactions on Asian Language Information Processing (TALIP), 8(4):14.
[25] Frank, Anette, Annie Zaenen, and Erhard Hinrichs. 2012. Treebanks: Linking linguistic theory to computational linguistics. Linguistic Issues in Language Technology, 7(1).
[26] Galley, Michel and Christopher D Manning. 2009. Quadratic-time dependency parsing for machine translation. In Proceedings of the Joint Conference of the 47th Annual Meeting of the ACL and the 4th International Joint Conference on Natural Language Processing of the AFNLP: Volume 2-Volume 2, pages 773–781, Association for Computational Linguistics.
[27] Gillenwater, Jennifer, Xiaodong He, Jianfeng Gao, and Li Deng. 2013. End-to-end learning of parsing models for information retrieval. In 2013 IEEE International Conference on Acoustics, Speech and Signal Processing, pages 3312–3316, IEEE.
[28] Habash, Nizar and Ryan M Roth. 2009. CATiB: The Columbia Arabic treebank. In Proceedings of the ACL-IJCNLP 2009 conference short papers, pages 221–224, Association for Computational Linguistics.
[29] Hajic, Jan, Otakar Smrz, Petr Zemánek, Jan Šnaidauf, and Emanuel Beška. 2004. Prague arabic dependency treebank: Development in data and tools. In Proc. Of the NEMLAR Intern. Conf. on Arabic Language Resources and Tools, pages110–117.
[30] Halabi, Dana, Arafat Awajan, and Ebaa Fayyoumi. 2017. Arabic lfg-inspired dependency treebank. In 2017 International Conference on New Trends in Computing Sciences (ICTCS), pages 207–215, IEEE.
[31] Han, Chung-hye, Na-Rae Han, Eon-Suk Ko, Martha Palmer, and Heejong Yi. 2002. Penn korean treebank: Development and evaluation. In Proceedings of the 16th Pacific Asia Conference on Language, Information and Computation, pages 69–78.
[32] Johan Hall, Jens Nilsson and Joakim Nivre. 2013. Malteval. http://www. maltparser.org/malteval.html.
[33] Katz-Brown, Jason, Slav Petrov, Ryan McDonald, Franz Och, David Talbot, Hiroshi Ichikawa, Masakazu Seno, and Hideto Kazawa. 2011. Training a parser for machine translation reordering. In Proceedings of the conference on empirical methods in natural language processing, pages 183–192, Association for Computational Linguistics.
[34] Kübler, Sandra, Ryan McDonald, and Joakim Nivre. 2009. Dependency parsing. Synthesis Lectures on Human Language Technologies, 1(1):1–127.
[35] Kulick, Seth, Ryan Gabbard, and Mitchell Marcus. 2006. Parsing the Arabic treebank: Analysis and improvements. In Proceedings of the Treebanks and Linguistic Theories Conference, pages 31–42.
[36] LDC. 2004a. Buckwalter Arabic morphological analyzer version. https://catalog.ldc.upenn.edu/LDC2004L02.
[37] LDC. 2004b. Prague Arabic dependency treebank 1.0. https://catalog.ldc.upenn.edu/docs/LDC2004T23/.
[38] LDC. 2007. Prague Arabic dependency treebank ++. http://padt-online.blogspot.com/2007/01/conll-shared-task-2007.html.
[39] LDC. 2018. 2007 CoNLL shared task – Arabic and English. https://catalog.ldc.upenn.edu/LDC2018T08.
[40] Li, Huiying and Feifei Xu. 2016. Question answering with dbpedia based on the dependency parser and entity-centric index. In 2016 International Conference on Computational Intelligence and Applications (ICCIA), pages 41–45, IEEE.
[41] Lynn, Teresa et al. 2016. Irish dependency treebanking and parsing.
[42] Maamouri, Mohamed, Ann Bies, Tim Buckwalter, and Wigdan Mekki. 2004. The Penn Arabic treebank: Building a large-scale annotated Arabic corpus. In NEMLAR conference on Arabic language resources and tools, volume 27, pages 466–467, Cairo.
[43] Maamouri, Mohamed, Ann Bies, Sondos Krouna, Fatma Gaddeche, and Basma Bouziri. 2009. Penn Arabic treebank guidelines. Linguistic Data Consortium. Marcus, Mitchell, Beatrice Santorini, and Mary Marcinkiewicz. 2006. Building a large annotated corpus of English: The Penn tree bank" in the distributed Penn tree bank project CD-ROM. Linguistic Data Consortium, University of Pennsylvania.
[44] McCaffery, Martin and Mark-Jan Nederhof. 2016. Dted: evaluation of machine translation structure using dependency parsing and tree edit distance. In Proceedings of the First Conference on Machine Translation: Volume 2, Shared Task Papers, volume 2, pages 491–498.
[45] Nilsson, Jens and Joakim Nivre. 2008. MaltEval: an evaluation and visualization tool for dependency parsing. In LREC. Nivre. 2018. MaltParser. http://www.maltparser.org/.
[46] Nivre, Joakim. 2005. Dependency grammar and dependency parsing. MSI report, 5133(1959):1–32.
[47] Nivre, Joakim. 2009. Parsing Indian languages with MaltParser. Proceedings of the ICON09 NLP Tools Contest: Indian Language Dependency Parsing, pages 12–18.

[48] Nivre, Joakim, Johan Hall, Sandra Kübler, Ryan McDonald, Jens Nilsson, Sebastian Riedel, and Deniz Yuret. 2007a. The conll 2007 shared task on dependency parsing. In Proceedings of the 2007 Joint Conference on Empirical Methods in Natural Language Processing and Computational Natural Language Learning (EMNLP-CoNLL).
[49] Nivre, Joakim, Johan Hall, and Jens Nilsson. 2006. Maltparser: A data-driven parser-generator for dependency parsing. In LREC, volume 6, pages 2216–2219.
[50] Nivre, Joakim, Johan Hall, Jens Nilsson, Atanas Chanev, Gül¸sen Eryigit, Sandra Kübler, Svetoslav Marinov, and Erwin Marsi. 2007b. Maltparser: A language-independent system for data-driven dependency parsing. Natural Language Engineering, 13(2):95–135.
[51] Owczarzak, Karolina, Josef Van Genabith, and AndyWay. 2007. Dependency-based automatic evaluation for machine translation. In Proceedings of the NAACL-HLT 2007/AMTA Workshop on Syntax and Structure in Statistical Translation, pages 80–87, Association for Computational Linguistics.
[52] Owens, Jonathan. 1988. The foundations of grammar: an introduction to medieval Arabic grammatical theory, volume 45. John Benjamins Publishing.
[53] Owens, Jonathan. 1990. Early Arabic grammatical theory: heterogeneity and standardization, volume 53. John Benjamins Publishing.
[54] Ryding, Karin C. 2005. A reference grammar of modern standard Arabic. Cambridge university press.
[55] Sima'an, Khalil, Alon Itai, YoadWinter, Alon Altman, and Noa Nativ. 2001. Building a tree-bank of modern hebrew text. Traitement Automatique des Langues, 42(2):247–380.
[56] Smrz, Otakar, Viktor Bielicky, and Jan Hajic. 2008. Prague Arabic dependency treebank: A word on the million words.
[57] Smrz, Otakar and Petr Pajas. 2004. MorphoTrees of Arabic and their annotation in the Tred environment. In Proceedings of the NEMLAR International Conference on Arabic Language Resources and Tools, pages 38–41.
[58] Smrz, Otakar, Jan Šnaidauf, and Petr Zemánek. 2002. Prague dependency treebank for arabic: Multi-level annotation of Arabic corpus. In Proc. of the Intern. Symposium on Processing of Arabic, pages 147–155.
[59] Solberg, Per Erik, Arne Skjærholt, Lilja Øvrelid, Kristin Hagen, and Janne Bondi Johannessen. 2014. The Norwegian dependency treebank.
[60] Tratz, Stephen C. 2016. Arl Arabic dependency treebank. Technical report, US Army Research Laboaratory Adelphi United States.
[61] Volk, Martin, Sofia Gustafson-Capková, David Hagstrand, and Heli Uibo. 2005. Teaching treebanking. Nordisk Sprogteknologi 2004: 2004: Aarbog for Nordisk Sprogteknologisk Forskningsprogram 2000-2004, page 143.
[62] Xia, Fei and Martha Palmer. 2001. Converting dependency structures to phrase structures. In Proceedings of the first international conference on Human language technology research, pages 1–5, Association for Computational Linguistics.
[63] Xue, Nianwen, Xiuhong Zhang, Zixin Jiang, Martha Palmer, Fei Xia, Fu-Dong Chiou, and Meiyu Chang. 2013. Chinese treebank 8.0 ldc2013t21. Linguistic Data Consortium, Philadelphia.
[64] Yu, Hui, Xiaofeng Wu, Wenbin Jiang, Qun Liu, and Shouxun Lin. 2015. An automatic machine translation evaluation metric based on dependency parsing model. arXiv preprint arXiv:1508.01996.